\documentclass[preprint,11pt,authoryear]{elsarticle}

\usepackage{multirow}
\usepackage{amsmath}
\usepackage{fontenc}
\usepackage{subfig}
\usepackage{bm}
\usepackage{amssymb}
\usepackage{graphicx}
\usepackage{multirow}
\usepackage{mathtools}
\DeclarePairedDelimiter{\norm}{\lVert}{\rVert}
\usepackage{graphicx}
\usepackage{stackengine}
\usepackage{wrapfig}
\usepackage{amssymb}
\usepackage[dvipsnames]{xcolor}
\usepackage{algorithm2e}
\DontPrintSemicolon
\SetNoFillComment
\usepackage{xcolor, soul}
\usepackage{array}
\newcolumntype{L}{>{\centering\arraybackslash}m{4.5cm}}
\newcolumntype{B}{>{\centering\arraybackslash}m{3.5cm}}
\newcolumntype{R}{>{\centering\arraybackslash}m{2.5cm}}
\newcolumntype{S}{>{\centering\arraybackslash}m{1.6cm}}

\def\bfm#1{{\bf #1}}

\def\bfs#1{\mbox{\boldmath{$ #1 $}}}

\newcommand\rd{d}

\definecolor{fgwhite}{rgb}{1,1,1}     
\definecolor{fgred}{rgb}{0.8,0,0}     
\definecolor{fgorange}{rgb}{0.93,0.53,0.18}     
\definecolor{fggreen}{rgb}{0,0.5,0}     
\definecolor{fgpurple}{rgb}{0.5,0,1}     
\definecolor{fggray}{rgb}{0.6,0.6,0.7}     
\definecolor{bggreen}{rgb}{0.8,1,0.8}     
\definecolor{fgblue}{rgb}{0,0,0.7}     
\definecolor{bgblue}{rgb}{0.9,0.9,1}     
\definecolor{fgclay}{rgb}{0.51,0.25,0.04}     

\sethlcolor{green}

\newcommand{\revised}[1]{{\color{black} #1}}

\usepackage{tikz}
\usetikzlibrary{tikzmark}

\setcitestyle{square}
\usepackage{lineno,hyperref}

\biboptions{authoryear}

\begin{document}

\begin{frontmatter}

\title{MAgNET: A Graph U-Net Architecture for Mesh-Based Simulations}






\author[mymainaddress]{Saurabh Deshpande}

\author[mymainaddress]{St\'ephane P.A. Bordas}

\author[mymainaddress,mysecondaryaddress]{Jakub Lengiewicz}


\address[mymainaddress]{Department of Engineering; Faculty of Science, Technology and Medicine; University of Luxembourg}
\address[mysecondaryaddress]{Institute of Fundamental Technological Research, Polish Academy of Sciences}

\begin{abstract} 
In many cutting-edge applications, high-fidelity computational models prove to be too slow for practical use and are therefore replaced by much faster surrogate models. Recently, deep learning techniques have increasingly been utilized to accelerate such predictions. To enable learning on large-dimensional and complex data, specific neural network architectures have been developed, including convolutional and graph neural networks. In this work, we present a novel encoder-decoder geometric deep learning framework called MAgNET, which extends the well-known convolutional neural networks to accommodate arbitrary graph-structured data. MAgNET consists of innovative Multichannel Aggregation (MAg) layers and graph pooling/unpooling layers, forming a graph U-Net architecture that is analogous to convolutional U-Nets. We demonstrate the predictive capabilities of MAgNET in surrogate modeling for non-linear finite element simulations in the mechanics of solids.  

\end{abstract}


\end{frontmatter}


\section{Introduction}

Computational models are essential tools for studying, designing, and controlling complex systems in many fields, including engineering, physics, biology, economics, and social networks. These models are often based on physical laws and mathematical equations, with partial differential equations (PDEs) being a common tool for describing how quantities change over space and time. In mechanics and physics, the PDEs are most commonly solved with numerical methods upon earlier space- and time- discretization, and a large number of domain-specific computational models have been developed so far, with the finite element method (FEM) and the finite volume method (FVM) being the most commonly used approaches in solid- and fluid mechanics, respectively. However, despite significant advances in computational performance over the last decade, such high-fidelity numerical simulations remain prohibitively expensive for many important applications, including emerging areas such as real-time feedback/control in the computer-assisted surgery \citep{johnsen2015niftysim, bui} or soft robotics \citep{rus2015design, goury2018fast}. Speeding up such models whilst maintaining the desired accuracy is an active area of research, and one of the main motivations of the present work.

Recently, deep learning (DL) techniques have taken a center stage across many disciplines. The DL models have proven to be accurate and efficient in predicting non-trivial nonlinear relationships in data. As such, they have been used for a variety of applications, including surrogate modeling in mechanics \citep{U-Mesh, NIKOLOPOULOS2022104652, DESHPANDE2022115307, vasilis, ZHANG2023106354, GLUMAC2023110135}, or model discovery and calibration \citep{HUANG2020109491,THAKOLKARAN2022105076}. The deep neural network approaches can be categorized with respect to how they use the data and \emph{a~priori} knowledge about the modelled system. In purely data-driven approaches, DL models rely on performing supervised learning on either experimental or numerically generated data and are agnostic to the underlying physics or model. As such, they are able to reproduce the physics-based relationship by implicitly learning on a relatively large amount of data \citep{FEM-deep, aydin2019general, daniel2020model, ML_CFD, HOQ2023106267}. If the \emph{a~priori} information about the modelled system is introduced, such networks are termed as Physics Informed Neural Networks (PINNs) \citep{PINN, SAMANIEGO2020112790, HENKES2022114790,  NGUYEN2022105176, ROY2023106049}. With respect to the purely data-driven approaches, PINNs are generally more accurate, require less data for training, and possess better generalization capabilities. The framework presented in the present work is generally applicable to both cases, however, for the sake of clarity, we will later only focus on purely data-driven types of networks. In any case, once trained, the DL models can be used as fast surrogates for computationally expensive high-fidelity numerical methods.

The focus of the present work is on high-dimensional relationships in which the sizes of inputs and/or outputs are large. Examples of such relationships can be found, for instance, in experimental full-field measurement data, such as our recent work on medical imaging \citep{LAVIGNE2022105490}, or in synthetic mesh data generated from finite element simulations, \citep{LORENTE2017342,PELLICERVALERO2020113083}. Although DL techniques have generally shown great success as efficient surrogates to computationally expensive numerical methods in scientific computing, some of the popular existing machine learning approaches are still based on fully-connected deep networks which are not suitable for high-dimensional inputs/outputs. As an alternative, the application of Convolutional Neural Networks (CNNs) has proven a promising performance in a wide variety of applications, also including accelerating non-linear finite element/volume/difference simulations \citep{CFD_CNN, cnnexample, DESHPANDE2022115307, ZHAO2023105516}. CNNs are designed to learn a set of fixed-size trainable local filters (convolutions), thus reducing the parameter space while being capable to capture non-linearities. In the context of computational mechanics, local convolutions leverage the natural local correlation of nearby nodes, which leads to more efficient neural network architectures, both in terms of training- and prediction times. Moreover, one can observe that the CNN architectures have a close analogy to some iterative solution schemes known in scientific computing \citep{unet_SR, multigrid_book}. This provides them with an additional interpretation of being trainable iterative computational schemes to solve sets of non-linear equations, rather than general-purpose black-box approximators.

However, there is one important limitation that prevents CNNs from being of general purpose. The problem is that they only work well with grid-like structure data, such as images or structured meshes, which greatly hinders their use for many real-world applications where data is structured differently. Although there are some attempts to alleviate that problem in the context of FEM data, for instance, combining finite elements with an immersed-boundary method \citep{immerserdboundary}, or embedding a precomputed coordinate mapping into the classic grid \citep{GAO2021110079}, the effectiveness of those methods is limited to simple irregular domains and remains challenging for complex geometries in general. A definitive solution to that problem has only been brought by Graph Neural Networks (GNNs)--architectures that directly handle arbitrarily-structured inputs/outputs. They belong to the recently emerged family of Geometric Deep Learning (GDL) methods which focus on neural networks that can learn from non-Euclidean input such as graphs and, more generally, manifolds \citep{GDL}\citep{graph_review}. Because of their ability to handle more general structured data, GNNs are gaining increasing importance also in surrogate modelling in scientific computing \citep{Sanchez-Gonzalez20208459, VLASSIS2020113299, pfaff2021learning, GAO2022114502} \revised{\citep{shivaditya2022graph}}\citep{SEO2023106284, JIANG2023106370, vasilisgraph}. However, these approaches are based on relatively simple message passing schemes, which are sub-optimal for learning on high non-linear regression tasks. In this work, we propose a novel local aggregation technique, which we denote as Multichannel Aggregation layer, MAg, which performs multichannel localised weighted aggregations, that can be seen as a direct extension of the traditional convolution layer in CNNs. Thanks to that, we are able to directly adapt some of the mechanisms/layers developed for CNNs to create efficient graph neural network architectures.

One mechanism that can improve the efficiency and predictive capabilities of convolutional and graph neural networks is the application of down-sampling (coarsening) and up-sampling (refinement) layers. In the context of CNNs, the focus is on encoder-decoder architecture frameworks, such as U-Net, which has been successfully implemented in various applications, including computer vision \citep{unet_original,3DUnet}, signal processing \citep{signalunet2,signalunet1}, and scientific machine learning \citep{U-Mesh,unet_nature,Pant_2021,su141911996, NIKOLOPOULOS2022104652,FERNANDEZLEON2023105945}. While the CNN-based U-Net approaches are limited to grid data, their graph-based version, known as graph U-Net, can provide the desired generality. Recently, various graph coarsening approaches have been proposed \citep{pool1,pool2,G-UNET,cai2021graph}, which serve the same function as pooling layers in CNNs, helping to reduce the size of a graph while maintaining essential properties of the processed data. In this work, we propose a novel graph pooling/unpooling operation (coarsening/refinement), that enables us to create a graph U-Net architecture, MAgNET, that can operate on arbitrary graphs. Our pooling layers are directly inspired by CNNs, where we extend the concept of pooling over local patches in regular grids to variable size non-overlapping cliques in graphs. This allows us to precompute coarsened graphs that are only based on the input graph topology, which is independent of data (i.e., node features). In the context of GNN-accelerated FEM simulations, a similar concept has been proposed by \citep{BLACK2022115120}, however, their implementation is limited to regular meshes for simple two-dimensional geometries and linear elastic problems.  Our approach enables computationally efficient deep learning models for non-linear problems involving arbitrary meshes, which is an important advancement for this field.

In summary, we introduce a novel graph U-Net framework comprising the proposed MAg and graph pooling/unpooling layers. The MAg layer captures local regularities in the input data, while the interleaved pooling layers reduce the graph representation to a smaller size while preserving important structural information. This enables us to efficiently implement our framework for large-scale problems. The proposed MAg and graph pooling layers are direct analogues of respective CNN U-Net layers and are also compatible with many state-of-the-art graph neural network layers. We elaborate on this point in the paper, providing a qualitative comparison of the proposed MAg layer with several existing graph layers. To validate the predictive capabilities of our framework, we apply it to several non-linear relationships obtained through finite element analysis and cross-validate it with predictions given by our CNN U-Net architecture \citep{DESHPANDE2022115307}. To increase the impact of our work, we provide source codes, datasets, and procedures to generate the datasets utilized in this work, which can be found in the MAgNET repository: \href{https://github.com/saurabhdeshpande93/MAgNET}{https://github.com/saurabhdeshpande93/MAgNET}.

The paper is organized as follows. In Section~\ref{sec: Methodology} we present the novel MAgNET framework, as well as its particular application to the hyperelastic FEM-based datasets. Then, in Section~\ref{sec: Results}, we provide details of implementation and a thorough study of MAgNET for several 2D and 3D benchmark non-linear FEM examples. The conclusions and future research directions are summarised in Section~\ref{sec: Conclusions}. 


\section{MAgNET Deep Learning Framework}
\label{sec: Methodology}

\begin{figure}[h]
     \centering
     \includegraphics[width=\textwidth]{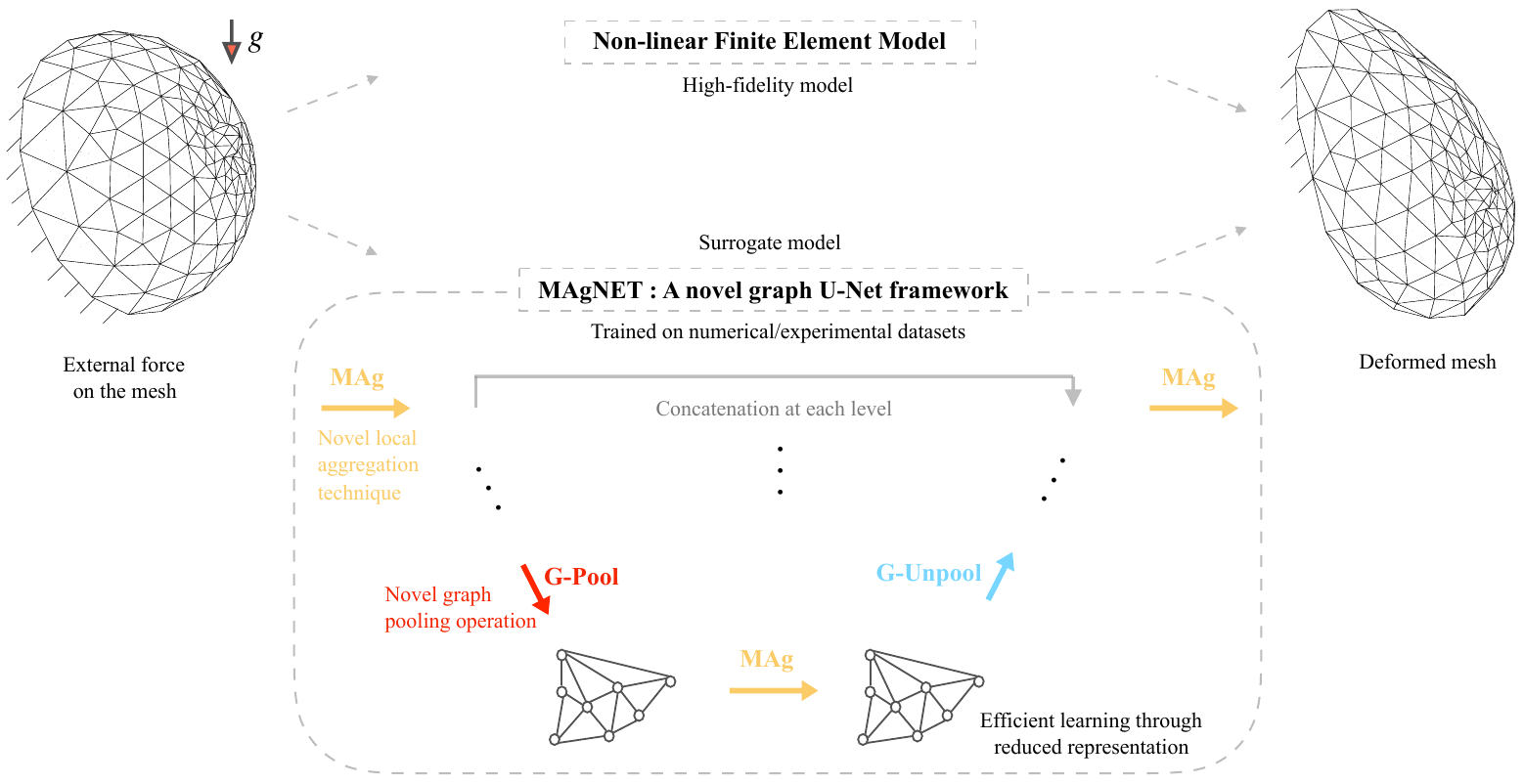}
     \caption{A novel graph U-Net neural network surrogate model for mesh-based simulations.
     MAgNET accurately captures non-linear FEM responses.}
     \label{fig: MAgNET schematic}
\end{figure}

In this section, we will propose a novel graph-based encoder-decoder (U-Net) deep-learning framework. We will provide a general formulation, in which inputs and outputs follow a certain graph topology (that is expressed by an adjacency matrix $\bfm{A}$). The graph expresses an assumed structure of correlations within input/output data and allows us to devise a \emph{robust} DNN architecture defining a non-linear mapping between inputs and outputs. We will apply this general framework to \emph{mesh-based} graphs. Such mesh topology of data is characteristic to spatially discretised numerical solution schemes for PDEs in scientific computing. In particular, we will focus on hyperelastic problems in solid mechanics, for which the training/testing data is obtained through the finite element method (see also the schematics in Figure~\ref{fig: MAgNET schematic}).


In Section~\ref{sec:graph_unet} we will provide an overview of the proposed Graph U-Net framework MAgNET. Next, in Sections~\ref{sec: Adjacency matrix}-\ref{sec: pooling_unpooling layer} we will introduce the building blocks of MAgNET. In particular, in Section~\ref{sec: Adjacency matrix}, we will introduce the adjacency matrix representation of the mesh-based graph, which will be utilised later in this paper; and in Sections~\ref{sec: Mag layer}-\ref{sec: pooling_unpooling layer} we will specify a new graph Multi-channel Aggregation (MAg) layer, as well as new graph pooling/unpooling layers. Afterwards, in Section~\ref{sec: information-passing interpretation}, we will provide an information-passing interpretation of the proposed Graph U-Net architecture. Finally, in Section~\ref{sec:fem_based_framework_intro} we will introduce a particular application of the framework to mesh-based datasets that are generated from FEM solutions of problems in hyper-elasticity.



\subsection{MAgNET architecture overview}
\label{sec:graph_unet}

\begin{figure}[h]
     \centering
     \includegraphics[width=\textwidth]{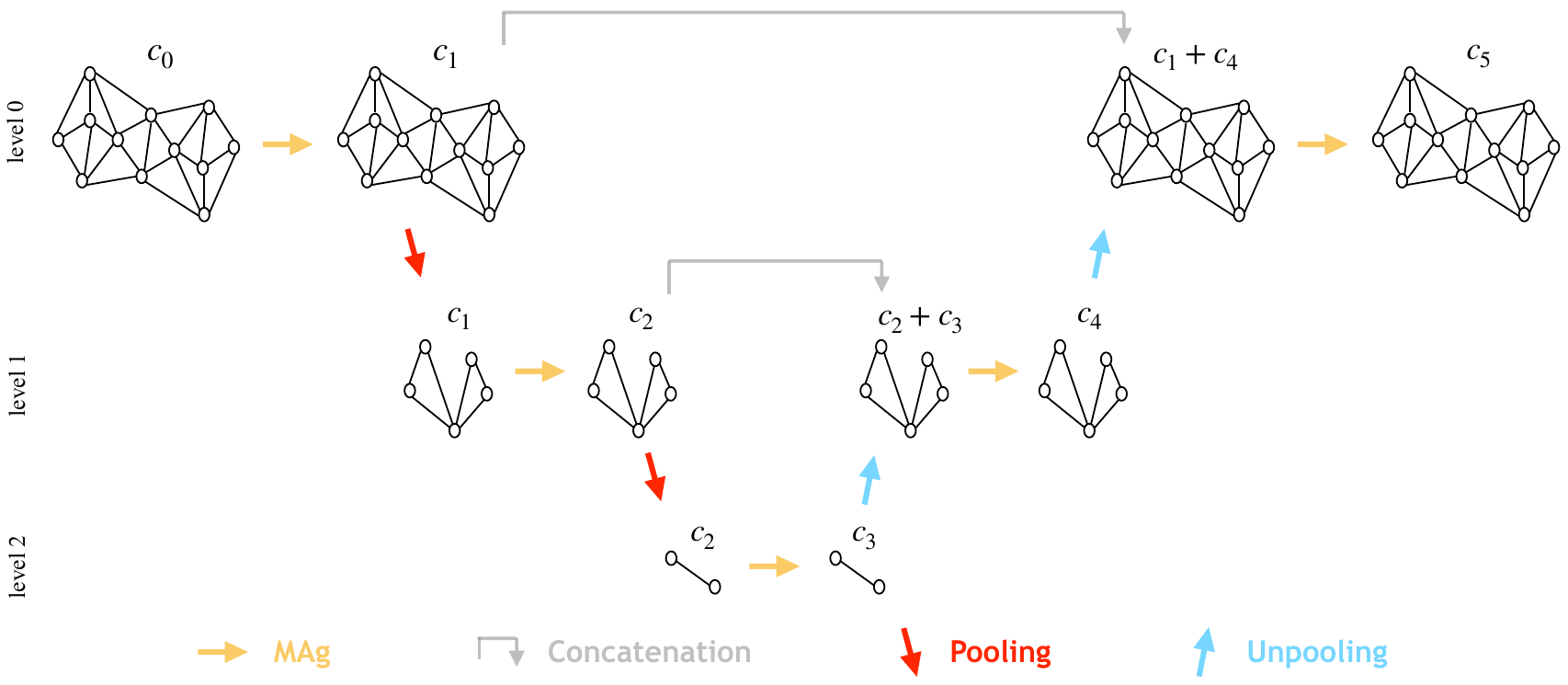}
     \caption{A schematic of Graph U-Net architecture for mesh based inputs. Colors indicate different types of layers. $c_1, c_2, \ldots, c_5$ stand for channel dimensions.
     Different arrows indicate different layers: the graph Multi-channel Aggregation (MAg) layer, the graph pooling/unpooling layers, and the concatenation layer.}
     \label{fig: Graph Unet}
\end{figure}

The MAgNET graph neural network architecture can be classified as a graph U-Net network and is an extension to the well-known class of convolution-based U-Net architectures (see~\citep{unet_original}). As such, the graph U-Net comprises of aggregation ('convolution'), pooling, unpooling, and concatenation layers (see the schematics in Figure~\ref{fig: Graph Unet}), which are here suitably adjusted to work with general (non-grid) topologies of inputs/outputs.



The graph U-Net architecture has two stages: encoding and decoding.
In the encoding stage, first, we apply one or more aggregation (MAg) layers, which are analogues of convolution layers in non-graph U-Net networks. Next, we apply a single graph pooling layer, which is a particular contraction of the graph, and which downsamples (coarsens) the problem. This aggregation-pooling sequence is repeated several times to achieve the desired level of contraction (coarsening). At the coarsest level, the MAg aggregation is performed one or more times, after which the decoding stage begins, which is the opposite to the encoding stage. At each level of decoding, the graph unpooling layer is followed by one or more MAg layers. At the upmost level, the last MAg layer is applied with linear activation to get the output.

More formally, the Graph U-Net network, $\mathcal{G}$, is constructed as follows.
First, we set the input layer $\bfs{d}^0$ as a vector of $N$ nodes, each of which being a vector of input values (also known as features or channels) of a constant length $c_0$. (Further on, we will refer to the features as the \emph{channels}.) Next, we subsequently add layers, $\bfs{d}^{l}$, to form a U-Net architecture. The subsequent layers, $\bfs{d}^{l-1}$ and $\bfs{d}^{l}$, are linked by the following relationship
\begin{equation}
    \bfs{d}^{l} = \textbf{T}^{l}(\bfs{d}^{l-1}; \bfs{\theta}^l),
\end{equation}
where $\bfs{\theta}^l$ is a vector of trainable parameters (e.g., weights and biases, $\bm{\theta}^l=\bfm{k}^l\cup\bfm{b}^l$), and $\textbf{T}^{l}(\cdot)$ is one of three already introduced transformations: MAg(), gPool() or gUnpool(), which will be more precisely defined in the following sections. Additionally, we also consider remote concatenation links between respective layers from the encoding and decoding stages, see Fig.~\ref{fig: Graph Unet}. The output layer, $\bfs{d}^L$, is assumed to be of the same mesh format as the in***********put layer but can have a different number of channels (features), $c_L$. Finally, we define the Graph U-Net network as a parameterized transformation
\begin{equation}
    \mathcal{G}(\bfs{d}^0,\bm{\theta})=\bfs{d}^L=\textbf{T}^{L}(\textbf{T}^{L-1}(\textbf{T}^{L-2}(\ldots); \bfs{\theta}^{L-1}); \bfs{\theta}^L),
    \label{eq: GraphUNet}
\end{equation}
where $\bfs{\theta}=\bigcup_{l=1}^L \{\bfs{\theta}^l\}$ is a concatenated vector of network parameters.






The calibration of the Graph U-Net parameters is done through a supervised learning, by fitting a given known input-output training dataset. 
The training dataset, 
\begin{equation}
\label{eq: dataset}
\mathcal{D}_{\text{tr}}=\{(\bfm{f}_1,\bfm{u}_1),...,(\bfm{f}_{M_{\text{tr}}},\bfm{u}_{M_{\text{tr}}})\},
\end{equation}
is in the mesh format, and the training is done by minimizing the following mean squared error loss function
\begin{equation}
      \mathcal{L}(\mathcal{D}_{\text{tr}},\bm{\theta}) = \frac{1}{M_{\text{tr}}}\sum_{m=1}^{M_{\text{tr}}} \norm{\mathcal{G}(\bfm{f}_m,\bm{\theta})-\bfm{u}_m}^{2} \label{eq:lossDeterm}
\end{equation}
which gives the optimal parameters  
\begin{equation}
      \bm{\theta}^{*} = \text{arg~} \underset{\bm{\theta}}{\text{min}}~\mathcal{L}(\mathcal{D}_{\text{tr}},\bm{\theta}). \label{eq:minimisationLossDeterm}
\end{equation}

\subsection{Adjacency matrix of the mesh-based graph}
\label{sec: Adjacency matrix}

\begin{figure}[ht]
     \centering
     \subfloat[]{
     \includegraphics[height=0.15\textwidth]{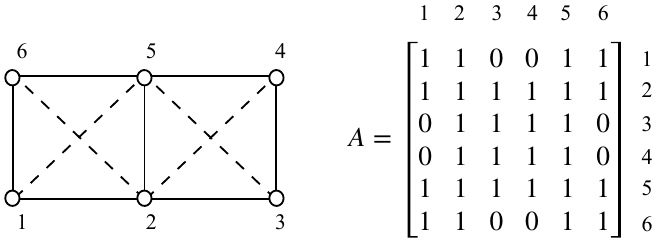}}\hspace{0.1\textwidth}
     \subfloat[]{
     \includegraphics[height=0.15\textwidth]{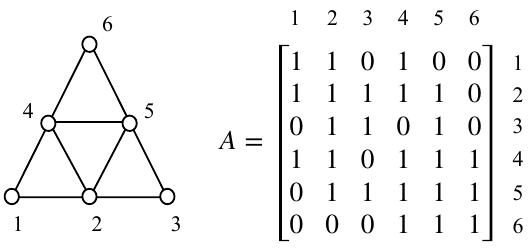}}
     \caption{Adjacency matrices for the (a) square and (b) triangular meshes. The dashed lines in (a) represent additional edges that are added to the original mesh.}
     \label{fig: adjacency matrices}
\end{figure}

For the purpose of this work, we will focus on sparse graphs that derive from data that is spatially organised in the form of meshes. Those can be 1D, 2D, or higher-dimensional meshes, of an arbitrary connection topology (see Fig.~\ref{fig: adjacency matrices} for examples of 2D meshes). The graph can be conveniently represented by a symmetric, square, Boolean adjacency matrix, $A$, whose order is equal to the number of nodes in the original mesh. To simplify the further notation, all nodes (vertices) are self-connected (have loops), which results in having $1$ on the diagonal of $\bfm{A}$. This allows us to more easily express certain graph operations that are used in this work, for instance, the k-th power of a graph $\bfm{A}$, and the selection of pooling sub-graphs that is presented in Section~\ref{sec: pooling_unpooling layer}.

It is fairly straightforward to generate an adjacency matrix from an element connectivity matrix of the mesh. For that reason we will not discuss it in detail. The only point to be emphasized is that we make all nodes belonging to a given element mutually inter-connected in the resulting graph (as they can be assumed to be strongly inter-related). We can visually represent it by adding more links as compared to a standard wire-frame visualization of finite-elements (see, e.g., the dashed lines in Fig.~\ref{fig: adjacency matrices}a).



\emph{Remark:} In our work we do not consider any attributes for the edges of a graph. Therefore, the data is only represented through nodal features and node-node connections which are defined through the adjacency matrix $\bfm{A}$.

\subsection{Multi-channel Aggregation (MAg) layer}
\label{sec: Mag layer}

The proposed novel neural network layer, MAg, is a multi-channel local aggregation layer that can operate on graph-structured data. Its architecture is a direct extension to the standard convolutional layer in CNNs, in which a shareable convolution window is used, making CNNs restricted to grid-structured data. In the MAg layer, instead, we propose to use fully-trainable local weighted aggregations (the so-called message passing scheme), where the aggregation neighborhood of a given node is prescribed through the graph connectivity (the adjacency matrix). As such, the scheme is very well suited for sparse graphs and can be directly applied to graphs that derive from arbitrary 2D or 3D meshes. 

\begin{figure}[h]
     \centering
     \subfloat[]{\includegraphics[width=0.37\textwidth]{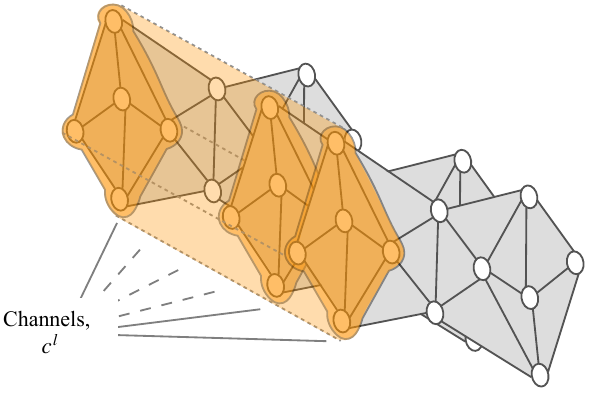}\label{MAg_schematic}}\hspace{0.1\textwidth}
     \subfloat[]{\includegraphics[width=0.32\textwidth]{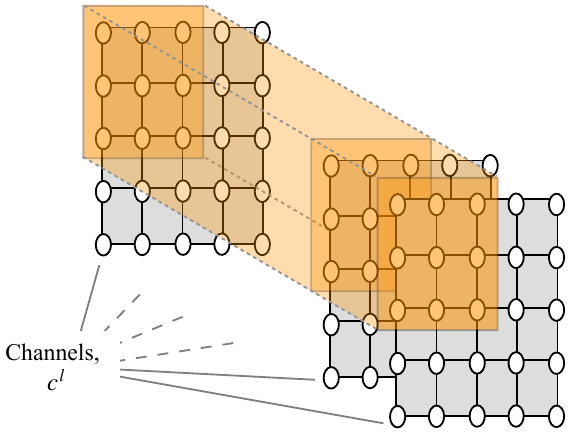}}
      \caption{Local aggregation in MAg (a) works very similar to the filter application in CNN (b). However as opposed to CNN, MAg uses different set of weights at different spatial locations with heterogeneous window size. In CNN, a constant filter slides across the channel.} 
     \label{fig: MAgvsCNN}
\end{figure}

The use of multiple channels aims to improve the capabilities of the network to capture non-linearities. In the multi-channel version, each node represents a vector of values (features), which can be visualised as multiple layers (channels) of the same graph structure (see the schematics in Figure~\ref{MAg_schematic}). The transformation between the input- and output multi-channel graphs is realised by applying multiple MAg aggregations on vector data to produce respective multiple components of output vectors. Note that the input/output channels of the whole network have usually a certain meaning, and their sizes are fixed (e.g., three RGB channels of a color image at the input and a single channel of a segmented image at the output). The number of channels in the latent layers can be chosen arbitrarily, which is up to the choice of a designer of a particular graph U-Net architecture.

More formally, we will consider the MAg layer as a parameterized transformation between the input and output nodes, defined as
\begin{equation}
    d^{l+1}_{i,\alpha} = \sigma( b^{l+1}_{i,\alpha} + \sum^{c^{l}}_{\beta = 1} \sum_{j\in \mathcal{N}_i} k^{l+1}_{i,j,\alpha,\beta}d^{l}_{j,\beta}),
    \label{eq: MAg}
\end{equation}
where $\mathcal{N}_i=\{j~|~A_{ij}=1\}$ is a set of neighbours of a node $i$ to be aggregated, $\alpha$ and $\beta$ represent the output and input channels, respectively, while $\bfm{k}^{l+1}$ and $\bfm{b}^{l+1}$ are trainable weights and biases, respectively. In this multi-channel definition, for a given component, $d^{l+1}_{i,\alpha}$, of an output, a single aggregation is performed throughout the neighborhood, $\mathcal{N}_i$, and all the input channels, $\beta\in\{1,\ldots,c^l\}$. The kernel parameters of MAg transformation, $k^{l+1}_{i,j,\alpha,\beta}$, are not shared, i.e, they can be independently trained for each aggregation window (note the free indexes $i$ and $\alpha$).

\subsubsection{Comparison to existing graph aggregation/convolution layers}


As already mentioned in the introduction, the very idea of generalisation of convolution layers to arbitrary graph structure is not new. In fact, various concepts have emerged so far, \citep{ZHOU202057}\citep{pmlr-v119-chen20v}, most of which are compatible with the U-Net framework proposed in the present work. Below, we will discuss several of them, introducing a unified notation that will facilitate a qualitative comparison with respect to the proposed MAg layer, see Table~\ref{tab: Graphconvs}.

\begin{table}[h!]
    \small
    \centering
    \begin{tabular}{l|c}
    Layer & Transformation \\[0.2em]
\hline 
    GCN \citep{kipf2016semi}& 
        \( \scriptsize
    d^{l+1}_{i,\alpha} = \sigma(\sum\limits^{c^{l}}_{\beta= 1} w^{l+1}_{\alpha,\beta}\sum\limits_{j\in \mathcal{N}_i} \frac{A_{i,j}}{\sum\limits_{k\in \mathcal{N}_i}A_{i,k}}  d^{l}_{j,\beta})
    \label{eq: GCN}
\) \\[0.5em]
    GAT \citep{GAT} & 
        \( \scriptsize
            d^{l+1}_{i,\alpha} =  \sigma(\sum\limits^{c^{l}}_{\beta = 1}w_{t,\gamma,\beta}^{l+1}\sum\limits_{j\in \mathcal{N}_i} \text{softmax}_{j}(\text{attn}(\bfs{w}_{t}^{l+1}\bfs{d}^{l}_{i},\bfs{w}_{t}^{l+1}\bfs{d}^{l}_{j},\bfs{\theta}_{\text{t}}))d^{l}_{j,\beta})
            \label{eq: GAT}
\) \\[0.5em]
    SemGCN \citep{Zhao_2019} & 
        \( \scriptsize
    d^{l+1}_{i,\alpha} = \sigma( \sum\limits^{c^{l}}_{\beta = 1}w^{l+1}_{\alpha,\beta} \sum\limits_{j\in \mathcal{N}_i} \text{softmax}_{j}(k^{l+1}_{i,j,\alpha,\beta})d^{l}_{j,\beta})
    \label{eq: SemGCN}
\)  \\
    MAg [present work] & 
        \( \scriptsize
    d^{l+1}_{i,\alpha} = \sigma(\sum\limits^{c^{l}}_{\beta = 1} \sum\limits_{j\in \mathcal{N}_i} k^{l+1}_{i,j,\alpha,\beta}d^{l}_{j,\beta})
    \label{eq: MAg tab} 
\)
    \end{tabular}
    \caption{Comparison of the MAg layer with selected state of the art graph convolution layers (biases are omitted for the sake of brevity). In GAT formulation, $\alpha=(t-1) N_{\text{g}} +\gamma$, which represents the stacking operation for the multi-head attention mechanism, where $t\in(1,\ldots,N_{\text{h}}^{l+1})$ is the attention head index, and $\gamma\in(1,\ldots,N_{\text{g}})$ is the internal channel index (cardinality of each node in the layer $l+1$). }
    \label{tab: Graphconvs}
\end{table}

Graph neural network layers aim to utilize the information about assumed correlations in data, with the graph structure expressing those correlations. The general approach is to specify a suitable (possibly nonlinear) trainable local transformation that can aggregate the information from a node in consideration and its neighbours. (This aggregation is followed by a chosen activation function before being propagated to the next layer.) Such transformations form a wide class of, so-called, message passing schemes, and can combine shareable (independent of a node) and non-shareable (dependent on a node, i.e., independently-trainable) sets of parameters. 


The simplest and most lightweight realisations of the graph aggregation/convolution layer concept only utilise shareable weights, see, e.g., the Graph Convolutional Network (GCN), \citep{kipf2016semi}. In those approaches, a non-trainable (arbitrary) weighted aggregation is performed prior to application of a shareable trainable operator -- something completely opposite to our MAg layer, which is fully trainable. This enables to keep the number of trainable parameters low, which is achieved at the cost of relatively low capacity of such networks. This low capacity can not be straightforwardly increased by simply deepening the network because of the well-known over-smoothing phenomenon.

We will discuss two out of many available approaches to increase the capacity of graph neural networks. The first approach relies on the multi-head attention mechanism which allows to assign different importance to nodes in the neighbourhood, see, e.g, the Graph Attention Network (GAT), \citep{GAT}. In the attention mechanism, the weights used in local aggregation depend on input nodal features, which makes the concept qualitatively different from all approaches (including the MAg layer) that use input-independent aggregation weights. The second class of approaches resemble the MAg layer more closely. Particular notable examples of that approach are the Spatial-Temporal Graph Convolution Network (ST-GCN), \citep{ST-GCN}, and the Semantic Graph Convolution Network (SemGCN), \citep{Zhao_2019}, which have been introduced in the particular context of human pose recognition problem (the computer vision domain). The common features of MAg and SemGCN layers are the input-independent learnable weighted aggregation and the use of channels to increase the model capacity. The difference is that the MAg doesn't use a shared transformation matrix ($\bfs{w}$) nor the softmax normalisation -- both used in the case of SemGCN.



To summarize, the proposed MAg layer relies on one of the most flexible message-passing schemes, with no shareable parameters. This promises a very high capacity of the MAg network. Also, as shown above, the proposed MAg layer is compatible with other graph convolution/aggregation layer concepts, and thus can be straightforwardly exchanged, if needed.

\subsection{Graph pooling- and unpooling layers}
\label{sec: pooling_unpooling layer}


Pooling and unpooling are two fundamental operations allowing U-Nets to encode (compress) and decode (decompress) information, respectively, see Figure~\ref{fig: Graph Unet}. The \emph{pooling} layers are composed of local contracting operations over the mesh-structured data, and are used to coarsen the data at the encoding part of the network. At the decoding part, the original refined mesh structure is restored by the \emph{unpooling} layers (upsampling operations). In U-Nets, the unpooling layer is usually combined with the \emph{concatenation} operation, which creates a direct link between the encoding and decoding part of the network (this will be explained later).

\begin{figure}[h]
     \centering
     \includegraphics[width=0.6\textwidth]{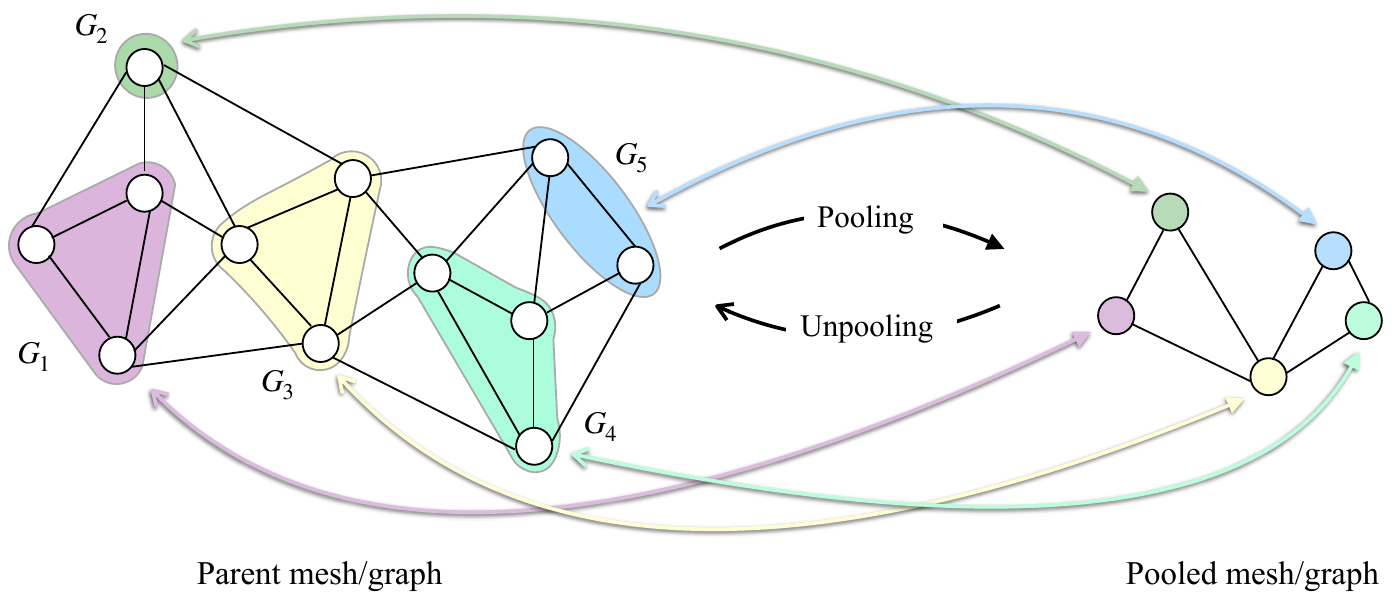}
     \caption{One arbitrary choice of non-overlapping subgraphs to create a pooled graph. Subgraphs $\bfm{G}_{1}, \ldots, \bfm{G}_{5}$ are represented with different colors and are generated by the Algorithm~\ref{alg:adjacency algo}.}
     \label{fig: graph pooling}
\end{figure}

\emph{Graph pooling}\newline
In this work, we propose a novel clustering-based graph pooling layer that can be applied to arbitrary graph-structured data. It can be seen as an extension to the pooling layers known from CNN U-Nets that are limited to grid-structured data. In our graph pooling approach, we split the graph into disjoint cliques (fully-connected subgraphs), and perform the contraction of all the identified cliques (i.e., every clique is replaced by a vertex, and new edges represent formerly connected cliques), see Figure~\ref{fig: graph pooling}. The split into cliques is done statically, i.e., at the graph U-Net construction phase. In particular, the split does not depend on the input data.

Below, we will provide a more formal construction of the pooling layer. For a given input graph $\bfm{G}$ that is represented by the vertices $\bfm{S}$ and the connectivity matrix $\bfm{A}$, we first generate an arbitrary set of $\tilde{N}$ non-overlapping fully-connected subgraphs (cliques) $\bfm{G}_1, \bfm{G}_2 ..., \bfm{G}_{\tilde{N}}$, i.e.,
\begin{equation}
    \bfm{S} = \bigcup^{\tilde{N}}_{i=1} {\bfm{S}_{i}}, \;\;\;\;\; \forall_{\bfm{S}_i} \forall_{j,k\in \bfm{S}_i}A_{jk}=1 \;\;\;\;\; \text{and} \;\;\;\;\; \forall_{i\neq j} \bfm{S}_i \cap \bfm{S}_j = \emptyset,
\end{equation}
where the sets $\bfm{S}_i$ represent nodes of the respective subgraphs $\bfm{G}_i$. The procedure to generate these subgraphs and the respective pooled adjacency matrix $\tilde{\bfm{A}}$ is described in Algorithm~\ref{alg:adjacency algo}. The pooled graph $\tilde{\bfm{G}}$ is composed of vertexes $\tilde{\bfm{S}}=\{1,\ldots,\tilde{N}\}$ with edges defined by the adjacency matrix $\tilde{\bfm{A}}$. The pooling layer is described as:
\begin{equation}
    d^{l+1}_{i,\beta}= \underset{j\in \bfm{S}_{i}}{\text{aggr}}~ d_{j,\beta}^{l}, 
\end{equation}
where the 'aggr' operation can be a max/min/avg, etc. Note that graph pooling layers do not modify the number of channels, i.e., the pooling is performed individually per each channel of the input.





\RestyleAlgo{ruled}
\SetKwComment{Comment}{/* }{ */}
\begin{algorithm}[h!]
\SetKwInput{KwData}{Input}
\caption{Generate a pooled graph from an arbitrary parent graph}\label{alg:adjacency algo}
\KwData{$N\times N$ adjacency matrix, $A$}
\KwResult{ list of subgraphs, $S$; \hspace{1em} $\tilde{N}\times \tilde{N}$ pooled adjacency matrix, $\tilde{A}$ }
\vspace{1mm}
$S \leftarrow$\{\} \Comment*[r]{initialisation of the subgraph list}
$P \leftarrow \{1,2,...,N\}$ \Comment*[r]{node indices of the parent graph}
$A' \leftarrow A$ \Comment*[r]{temporary copy of matrix $A$}
\vspace{1mm}
\break
\Comment{Loop for generating non-overlapping subgraphs, $S$, see Figure~\ref{fig: graph pooling}} 
\break
 \While{\text{P} $\neq$ null}{
  $p\in P$ \Comment*[r]{randomly select a single node} 
  $S_i \leftarrow \{p\}$\Comment*[r]{initialise subgraph}
  $\mathcal{N}_p \leftarrow \{m\neq p~|~A'[m,p]=1\} $\Comment*[r]{nodes connected to selected node}
  \For{$n$ in $\mathcal{N}_p$}{
   \If{ $\forall~m\in S_i \;\; A'[m,n]=1$}{
   $S_i\leftarrow S_i \cup \{n\}$ \Comment*[r]{append node to the subgraph}
   }
   }
   {
   $P \leftarrow P \setminus S_i$ \Comment*[r]{remove subgraph from parent graph}
    $\forall~m\in S_i \;\; A'[m,:] \leftarrow 0; \, A'[:,m] \leftarrow 0$ \Comment*[r]{remove subgraph from parent graph} 
    $S \leftarrow S \cup S_i$ \Comment*[r]{add subgraph to subgraphs list}
  }
 }
 \vspace{1mm}
\break
$\tilde{N}=\text{sizeof}(S)$\Comment*[r]{number of pooled nodes = number of pooling subgraphs}
$\tilde{A}\leftarrow$ zeros($\tilde{N},\tilde{N}$)\Comment*[r]{zero initialisation of pooled matrix}
\tcc{Loop for constructing pooled adjacency matrix $\tilde{A}$ from subgraphs $S$} 
\break
 \For{$r$ in $\{1,2,..,\tilde{N}\}$}{
  \For{$c$ in $\{1,2,..,\tilde{N}\}$}{
   \If{ $\exists ~ n\in S[r]$, $m\in S[c]~|~ A[n,m]=1$}{
   $\tilde{A}[r,c]\leftarrow 1$  \Comment*[r]{if $S[r]$ and $S[c]$ are connected by an edge}}}}
\end{algorithm}


Graph pooling can be applied several times at the encoding part of the U-Net, e.g., see Fig.~\ref{fig: Graph Unet}. For the purpose of future unpooling operations, after each pooling operation, we save the original graph, $\bfm{G}$, the adjacency matrix, $\bfm{A}$, and the pooling subgraphs, $\bfm{G}_i$. After doing so, we substitute $\bfm{G}\leftarrow\tilde{\bfm{G}}$, and $\bfm{A}\leftarrow\tilde{\bfm{A}}$.

\emph{Graph unpooling}\newline
Structure-wise, the graph unpooling is a reverse operation to pooling. More precisely, the output graph of an unpooling layer will have the same topology as the input graph of the related pooling layer, see Figure~\ref{fig: graph pooling}. The operation is defined via the previously saved subgraphs $\bfm{G}_j$ (with nodes $\bfm{S}_j$) as
\begin{equation}
    d^{l+1}_{i,\beta}=  d_{j,\beta}^{l}\hspace{2em}\text{for}\hspace{1em}i\in \bfm{S}_{j},
    \label{eq: unpooling}
\end{equation}
and it simply replicates the features of a node $j$ to the nodes specified by $\bfm{S}_j$. As such, this operation is analogous to the related upsampling operation used in CNNs.


\emph{Graph unpooling + concatenation}\newline
Concatenations, also known as skipped connections, are characteristic to U-Net architectures. Thanks to them, the layers from the decoder part gain a direct access to features from the encoder part. Concatenations help to mitigate the issue of vanishing gradients, and add extra information that could have been lost due to the earlier downsampling (pooling). 

In our case, the concatenations are always related to the respective pooling/unpooling operation pairs, see Fig.~\ref{fig: Graph Unet}. It is done by stacking the output of an unpooling layer $l$, given by Equation~(\ref{eq: unpooling}), with the input of a respective pooling layer $l'$:
\begin{equation}
     d^{l+1}_{i,c^{l}+\alpha}= d^{l'}_{i,\alpha},
\label{eq:concatenation}
\end{equation}
In the formula above, $c^{l}$ is the number of channels in unpooling inputs. As the result, the total number of output channels of unpooling+concatenation is $c^{l}+c^{l'}$.

\subsection{Information-passing interpretation of MAg and pooling layers}
\label{sec: information-passing interpretation}

\begin{figure}[h]
     \centering
     \includegraphics[width=\textwidth]{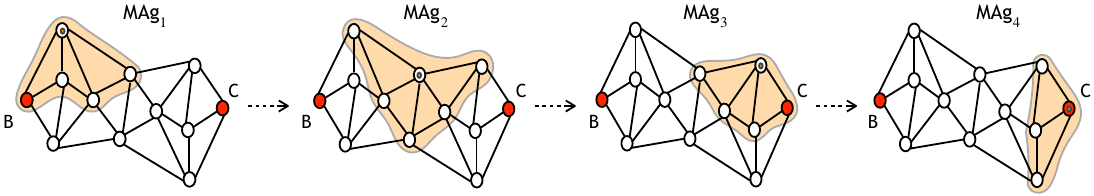}
     \caption{This 2D mesh requires at least 4 subsequent local aggregation operations (orange areas with center nodes marked by dots) to propagate the feature information from node B to the distant node C.}
     \label{fig: Information passage}
\end{figure}

During a single forward pass of the MAg layer, the aggregation is performed locally for each individual node, i.e., each node of the graph will have an access to the aggregated feature information from its adjacent nodes only, specified by the adjacency matrix, $\bfm{A}$, see Eq~(\ref{eq: MAg}). Therefore, the nodes that are not directly connected through $\bfm{A}$ do not exchange information at a single MAg operation (see Figure~\ref{fig: Information passage}). Such long-distance exchange across the network is fundamental to allow the neural network model to express correlations between topologically distant input- and output nodes (e.g., how the output displacements at node C depend on the input loads at the node B, in Figure~\ref{fig: Information passage}). 

One way to handle this issue would be to apply the MAg layer several times as shown in Figure~\ref{fig: Information passage}. However, in that case, the number of subsequent layers would be proportional to the diameter of the underlying graph, which could increase the number of training variables and the depth of the network, deteriorating its performance. A natural simple improvement, also utilised in the present paper, is to increase the support (neighbourhood) of the MAg operations. In the proposed framework, this can be straightforwardly done by using higher powers of the adjacency matrix, e.g., $\bfm{A}^{\!2}$ or $\bfm{A}^{\!3}$, instead of $\bfm{A}$. This improvement alone, however, would still require the number of MAg layers to be proportional to the graph diameter. 

\begin{figure}[h]
     \centering
     \includegraphics[width=0.65\textwidth]{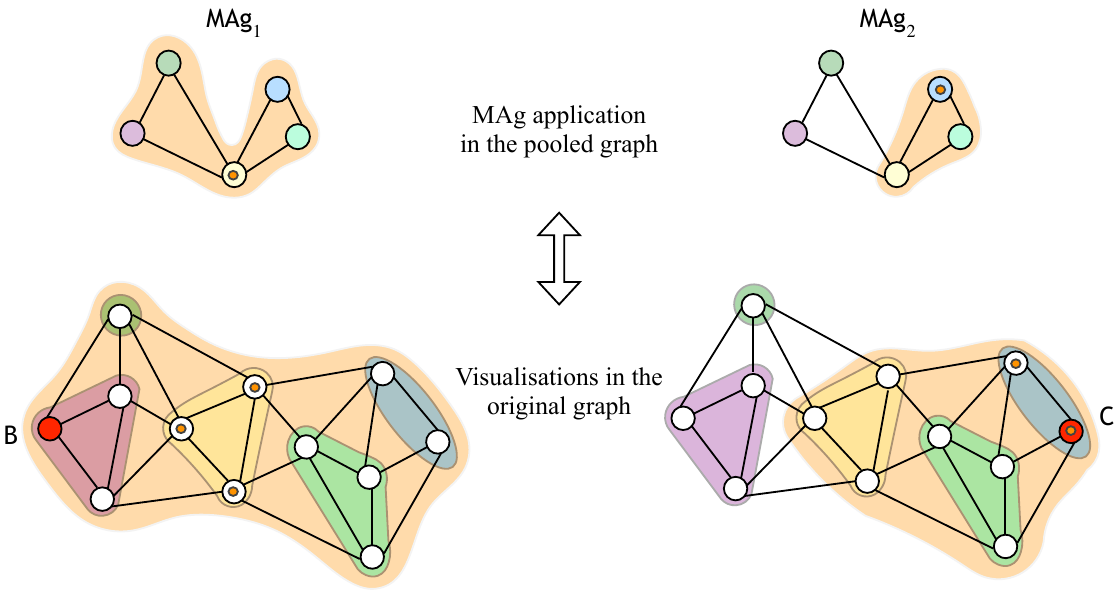}
     \caption{Visualisation of feature information exchange between nodes in the pooled graph. In the pooled space, only 2 MAg operations are sufficient to exchange feature information between spatially further located nodes in the original graph. The orange region shows the window of MAg operation.
     }
     \label{fig: pool viz}
\end{figure}

The above observations explain a natural motivation behind using the pooling/unpooling layers, and hence creating the U-Net architecture. The pooled graph can be seen as a reduced space representation of the parent graph, and each pooled node aggregates the feature information corresponding to multiple nodes of the parent graph, see Figure~\ref{fig: pool viz}. The pooled graph is of a coarsened topology when compared to the parent graph, and this allows for the feature information exchange with a lower number of MAg layers. The pooling/unpooling layers can be nested, which provides an exponential reduction rate of the graphs' diameters.

\emph{Remark:} Note that the pooling layer proposed in this work represents a clique-pooling approach in which the cliques are non-overlapping. This allows us to achieve a very good level of graph coarsening (contraction). It is unlike a similar clique-based strategy that has been recently proposed, \citep{cliquepool}, in which the pooling cliques overlap, providing a much lower level of coarsening.

\subsection{Application to FEM-based datasets} 
\label{sec:fem_based_framework_intro}

We will now focus on a particular graph structure of inputs/outputs that will be in a form of finite element mesh. Specifically, the MAgNET framework will be applied as a surrogate model to a finite element model in large-deformation elasticity. The finite element model will be shortly introduced below.

We consider a boundary value problem expressed in the weak form over the domain $\Omega$:
\vspace{5mm}
  \begin{equation}
    \label{eq:VWP}
          \int_{\Omega} \bfs{P}(\bfs{F}(\bfs{u})) \cdot \nabla\delta\bfs{u}\, \rd V - \int_{\Omega} \rho\,\bfs{\bar{b}} \cdot \delta\bfs{u}\, \rd V -
      \int_{\Gamma_t} \bfs{\bar{t}} \cdot \delta\bfs{u}\, \rd S   = 0 \quad\quad\forall\delta\bfs{u},
  \end{equation}
where $\bfs{P}(\bullet)$ is the first Piola-Kirchhoff stress tensor, $\bfs{\bar{b}}$ are prescribed body forces, $\bfs{\bar{t}}$ are prescribed tractions on the Neumann's boundary $\Gamma_N$, while the solution $\bfs{u}$ and the variation $\delta\bfs{u}$ belong to appropriate functional spaces, with $\bfs{u}=\bar{\bfs{u}}$ and $\delta\bfs{u}=\bfm{0}$ on the Dirichlet boundary $\Gamma_u$. The hyperelastic constitutive relationship is expressed through the strain-energy density potential $W(\bfs{F})$ as
  \begin{equation}
    \label{eq:constitutive_general}
    \bfs{P}(\bfs{F}) = \frac{\partial W(\bfs{F})}{\partial \bfs{F}},
  \end{equation}
where the deformation gradient tensor $\bfs{F}=\bfm{I} + \nabla \bfs{u}$.

For all the cases considered in the present work, the Neo-Hookean hyperelastic law with the following strain energy potential is used, see~\citet{simo1982penalty}, 
\begin{equation}
 W(\bfs{F})=\frac{\mu}{2}(I_c-3-2 \ln{J})+\frac{\lambda}{4}( J^2-1-2 \ln{J}),
 \label{Eq:NeoHoohe}
\end{equation}
where the invariants $J$ and $I_{\text{c}}$ are given in terms of deformation gradient $\bfs{F}$ as
\begin{equation}
    J = \text{det}(\bfs{F}), \quad
    I_{\text{c}} = \text{tr}(\bfs{F}^{T}\bfs{F}), \quad,
\end{equation}
with $\mu$ and $\lambda$ being Lame's parameters, which can be expressed in terms of Young's modulus, $E$, and Poisson's ratio, $\nu$, as
\begin{equation}
    \lambda = \frac{E\nu}{(1+\nu)(1-2\nu)}, \quad \mu = \frac{E}{2(1+\nu)}.
\end{equation}
Note that one can use other forms of the volumetric part of the above potential, see~\citet{doll2000development}, or other hyperelastic models, such as the Mooney-Rivlin and a more general class of Ogden models, see~\citet{ogden2005}.



Finite element discretisation transforms the weak form expressed by Eq.(\ref{eq:VWP}) into the system of non-linear equations
\begin{equation}
    \label{eq:res}
        \mathbf{R}(\mathbf{u}; \mathbf{f}_{\text{ext}})=\mathbf{f}_{\text{int}}(\bfm{u})-\mathbf{f}_{\text{ext}}=\bfm{0}, 
 \end{equation}
 that expresses the balance between external and internal nodal forces. In this work, the vector of external forces, $\mathbf{f}_{\text{ext}}$, represents boundary conditions, which can be surface tractions or body forces. Given $\mathbf{f}_{\text{ext}}=\mathbf{f}_m$, the system is solved for an unknown vector $\mathbf{u}$ with the Newton-Raphson scheme, giving as a result the solution $\mathbf{u}_m$. A pair $(\mathbf{f}_m, \mathbf{u}_m)$ makes an element of the dataset $\mathcal{D}$ introduced in Eq.~(\ref{eq: dataset}), and the FE mesh that results from the FE discretization produces the adjacency matrix $\bfs{A}$ introduced in Section~\ref{sec: Adjacency matrix}.
\section{Results}
\label{sec: Results}

In this section, we study the performance of the proposed framework in application to surrogate modeling in mechanics of solids. For that purpose, we use four benchmark problems. In Section~\ref{sec: data_generation_FEM}, we give a detailed specification of the benchmark problems and the procedure for obtaining FEM-based datasets. In Section~\ref{sec: NN_implementation_and_training}, we provide details of neural network architectures for each of the studied cases and will describe the training procedure. In Section~\ref{sec: cross_validation_of_NN_models}, we study the predictions of neural network models by cross-validating results from MAgNET and CNN models, and by comparing them with the FEM results. Finally, in Section~\ref{sec: predictions_MAgNET_unstructured} we demonstrate the capabilities of the MAgNET framework to provide a surrogate model for the unstructured mesh cases.

\subsection{Generation of FEM based datasets}\label{sec: data_generation_FEM}

\begin{figure}[ht!]
     \centering
      \subfloat[2D L-shape]{\stackinset{r}{0.29\textwidth}{t}{0.045\textwidth}{\includegraphics[width=0.05\textwidth]{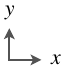}}{\includegraphics[width=0.37\textwidth]{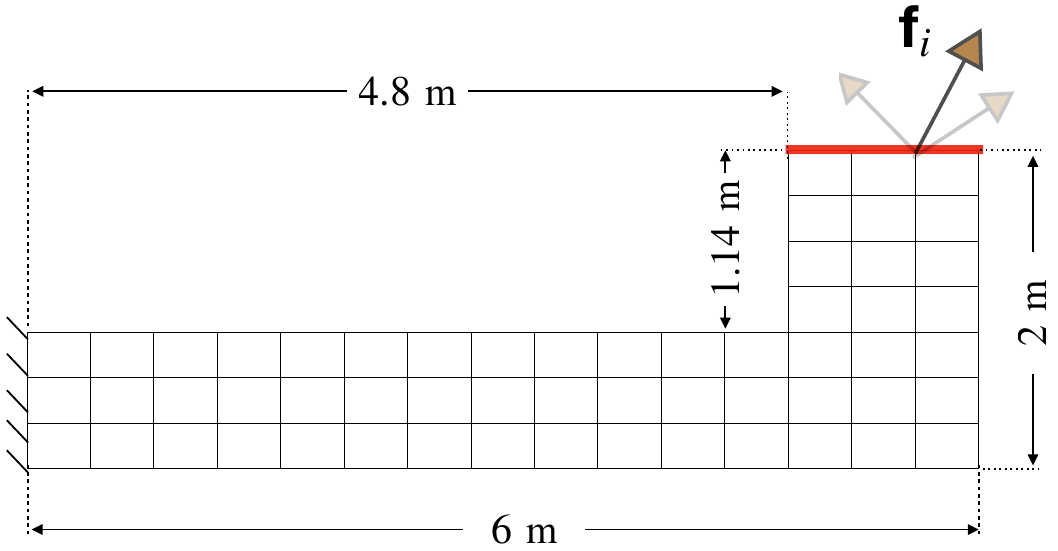}\label{L_shape}}}\hspace{0.09\textwidth}
      \subfloat[3D beam]{\stackinset{r}{0.04\textwidth}{t}{0.022\textwidth}{\includegraphics[width=0.05\textwidth]{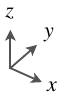}}{\includegraphics[width=0.37\textwidth]{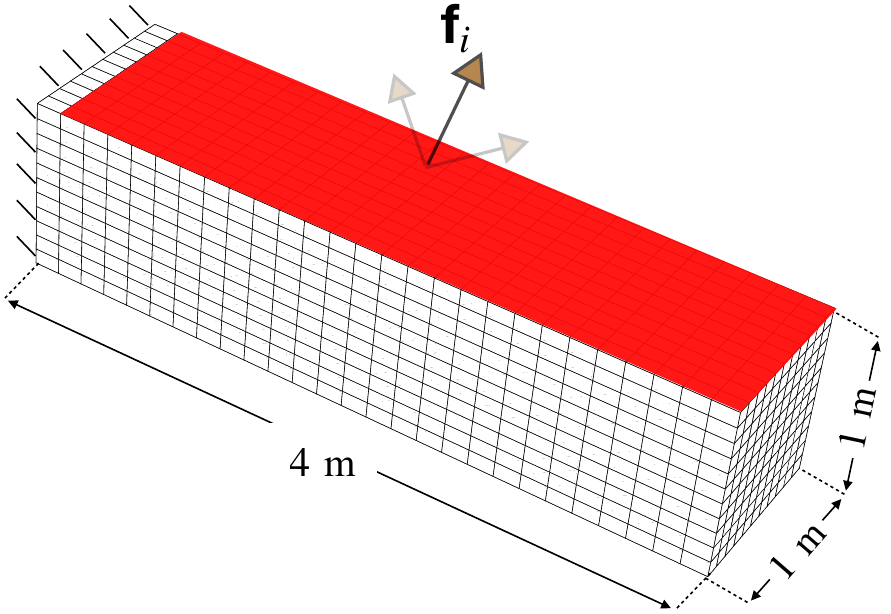}\label{coarse_beam}}}
    
      \subfloat[2D beam with hole]{\stackinset{r}{0.25\textwidth}{t}{-0.026\textwidth}{\includegraphics[width=0.05\textwidth]{Images/axis_2d.pdf}}{\includegraphics[width=0.34\textwidth]{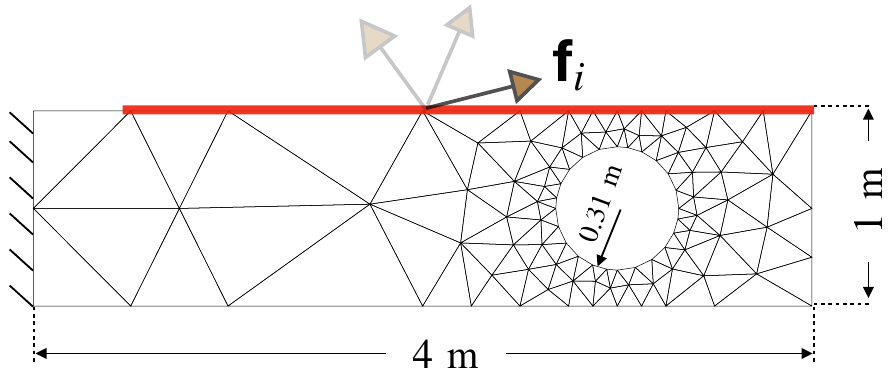}\label{2d_phole}}}
      \hspace{0.15\textwidth}
     \subfloat[3D breast]{\stackinset{r}{-0.08\textwidth}{t}{0.08\textwidth}{\includegraphics[width=0.06\textwidth]{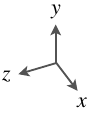}}{\includegraphics[width=0.24\textwidth]{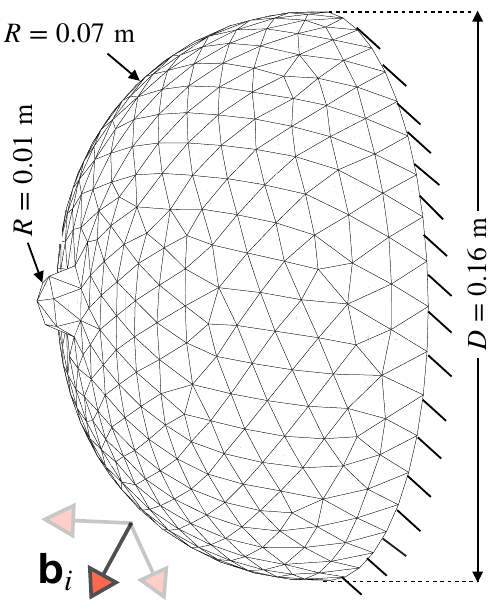}\label{3d_breast}}}
    
     \caption{Schematics of four benchmark problems. (a) 2D L-shape geometry (quad mesh), (b) 3D beam geometry (hexahedron mesh), (c) 2D beam with hole geometry (triangular mesh), and (d) 3D breast geometry (tetrahedron mesh). In examples (a)-(c), single nodal loads are applied on the region of boundary indicated with red color. In example (d), only body forces are considered.}
     \label{fig: benchmark examples schematics}
\end{figure}

We consider four benchmark problems, see Figure~\ref{fig: benchmark examples schematics}. Two of them,  Figure~\ref{fig: benchmark examples schematics}(a-b), utilise simple meshes, which makes it possible to assure structured (grid) inputs. They will be used to cross-validate between our MAgNET architecture and the standard CNN U-Net architecture. The other two examples, Figure~\ref{fig: benchmark examples schematics}(c-d), are more complex and will serve us to demonstrate the applicability of MAgNET for general (unstructured) meshes. Each of those two groups consists of a 2D and a 3D problem, thanks to which the framework can be tested for four different finite element topologies: triangular, quadrilateral, tetrahedral, and hexahedral.

For all considered cases, we use the neo-Hookean material model, see Section~\ref{sec:fem_based_framework_intro}, with material parameters provided in the Table~\ref{tab:material_properties}. In order to generate training/testing datasets, for each discretized problem we individually specify a family of boundary conditions, as described below, see also schematics in Figure~\ref{fig: benchmark examples schematics}. For the cases shown in Figures~\ref{fig: benchmark examples schematics} (a-c), nodes on one side are fixed (Dirichlet boundary conditions), and only a single random nodal force is applied at a selected node in a prescribed region of interest (denoted by red line/surface in Figure~\ref{fig: benchmark examples schematics}(a-c)). For the remaining nodes, the external forces are set to $\bfm{0}$. For the case shown in Figure~\ref{fig: benchmark examples schematics}(d), the uniform body force density is prescribed (force per unit mass, with density $\rho=1000~\text{kg}/\text{m}^3$). The body force field is integrated through element shape functions to obtain respective nodal forces that are used in datasets.

\begin{table}[h!]
    \small
    \centering
    \begin{tabular}{ll|c|R|S|S}
        & Problem (element topology) & Is structured? & Young's modulus, $E$ [\text{Pa}] & Poisson's ratio, $\nu$ & Density, $\rho$ [kg/m$^3$] \\
        \hline
        a) & 2D L-shape (quad) & No (Yes) & 500 & 0.4 & - \\
        b) & 3D beam (hexahedron) & Yes & 500 & 0.4 & - \\ 
        c) & 2D beam with hole (triangular) & No & 500  & 0.3 & - \\ 
        d) & 3D breast (tetrahedron) & No & 800 & 0.4 & 1000
    \end{tabular}
    \caption{Material properties used for the benchmark cases.}
    \label{tab:material_properties}
\end{table}

All the finite element computations were implemented and performed within the AceGen/AceFem framework~\citep{acegen}. For a given problem, for each loading case, $i$, the entire vector $\bfm{f}_{(i)}$ of external nodal forces and the vector $\bfm{u}_{(i)}$ of computed nodal displacements were saved, which allowed to generate the final training/testing dataset $D=\{(\bfm{f}_{(1)},\bfm{u}_{(1)}),...,(\bfm{f}_{(M_{\text{tr}}+M_{\text{te}})},\bfm{u}_{(M_{\text{tr}}+M_{\text{te}})})\}$. The datasets were randomly split into training sets, $M_{\text{tr}}$ (95\%), and testing sets, $M_{\text{te}}$ (5\%). The sizes of datasets and the distribution of force magnitudes are provided in Table~\ref{tab:datasets}.




\begin{table}[h!]
    \small
    \centering
    \begin{tabular}{ll|S|L|R|S}
        & Problem & N.of FEM DOFs ($\mathcal{F})$ & Range (External forces/ body force density) & Dataset size $M_{\text{tr}}+M_{\text{te}}$ & samples per~node  \\
        \hline
        a) & 2D L-shape & 160 & $f_x, f_y$ = -1 to 1 N & $3800 + 200$ & 1000\\
        b) & 3D beam & 12096 & $f_x, f_y, f_z$= -2 to 2 N & $33858 + 1782$ & 110 \\ 
        c) & 2D beam (hole) & 198 & $f_x, f_y$ =  -5 to 5 N  & $4560 + 240$ & 400\\
        d) & 3D breast & 3105 & $b_x, b_y$ = -6 to 6 $\text{N}/\text{kg}$ , \hspace{9mm}$b_z$ = -3 to 3 $\text{N}/\text{kg}$ & $7600 + 400$ & -
    \end{tabular}
    \caption{Specification of FE-based datasets. For cases (a-c), the external force is applied at a selected node, and for case (d), external body forces are applied. The magnitudes of forces are randomly sampled from the multivariate uniform distribution, with ranges specified in the table. For cases (a-c), multiple samples per node are generated, for all nodes in the prescribed area of interest.}
    \label{tab:datasets}
\end{table}



\subsection{Design, implementation and training of neural network models}\label{sec: NN_implementation_and_training}

The implementation of the layers and mechanisms of the MAgNET framework described in Section~\ref{sec: Methodology} and of CNN U-Net framework introduced in \citep{DESHPANDE2022115307} has been performed within the TensorFlow libraries. We use them to build and train deep neural network models for the cases described in Section~\ref{sec: data_generation_FEM}. Table~\ref{tab: architectures} outlines individual properties of the network architectures implemented in this work. The codes and datasets are publicly available open source \citep{DeshpandeMAgCode2023}, which makes it possible for other researchers to reproduce the present results and also to extend our frameworks to new cases/problems.

\begin{table}[h]
\small
\begin{center}
 \begin{tabular}{l| c | B | B | S} 
 Example & Network type & (N. of poolings, N. of MAg/conv. layers per~level, window size) & N. of channels per~level & N. of parameters \\ [0.7ex] 
 \hline
\multirow{2}{*}{2D L-shape} & MAgNET & (3, 2, $A^2$) & 16, 32, 64, 128 & \multirow{2}{*}{$\sim$ 4 E+6}   \\ 
 & CNN U-Net & (2, 2, $3\!\times\!3$) & 64, 128, 512  &  \\
 \hline
\multirow{2}{*}{3D beam} & MAgNET & (5, 1, $A^2$)  & 3, 3, 3, 12, 24, 48 & \multirow{2}{*}{$\sim$ 75 E+6}\\
   & CNN U-Net & (4, 2, $3\!\times\!3\!\times\!3$) & 256, 256, 256, 512, 512 &  \\
\hline
 2D beam (hole) &MAgNET & (3, 2, $A^2$) & 8, 16, 32, 64 & $\sim$ 2 E+6 \\
\hline
3D breast & MAgNET & (4, 1, $A^2$) & 6, 12, 12, 24, 48  & $\sim$ 19 E+6
\end{tabular}
\end{center}
\caption{Neural network architectures implemented in this work. The leaky ReLU activation function is used in all MAgNET cases, while ReLU activation is used for CNN cases. For the last layers, the linear activation function is always applied.}
\label{tab: architectures}
\end{table}

\begin{figure}
    \centering
    \includegraphics[width=0.95\textwidth]{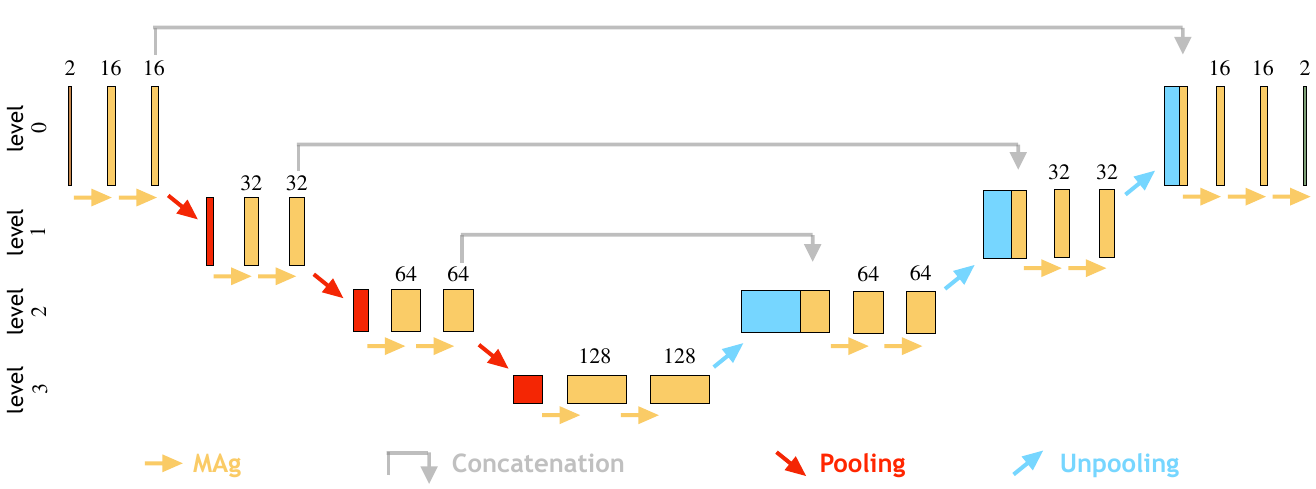}
    \caption{MAgNET architecture used for the 2D L-shape example.}
    \label{fig:lshapearchitecture}
\end{figure}

To provide a complete understanding of the neural network architectures listed in Table~\ref{tab: architectures}, we will now delve into the details of the MAgNET architecture used for the 2D L-shape example. Its schematics is shown in Figure~\ref{fig:lshapearchitecture}. As indicated in the third column in Table~\ref{tab: architectures}, it is a three-level graph U-Net architecture with two MAg operations at each level. The fourth column specifies the number of channels utilized for the MAg operations at each level of the graph U-Net. The forward pass starts with the input mesh (2D), to which the MAg layer is applied twice (with 16 output channels). This is followed by the graph pooling layer, which coarsens the mesh and transitions to the next level of the U-Net (from zeroth to the first level). This process repeats twice, with the first and second levels of the graph U-Net having MAg layers with 32 and 64 output channels, respectively, leading to the coarsest third level of the U-Net. At this level, two MAg layers (with 128 output channels) are applied. In the subsequent decoding phase, the graph unpooling layer is employed with the concurrent concatenation operation, and followed by two MAg operations (with 64 output channels). This upsampling sequence repeats twice with the use of 32- and 16-channel MAg layers. Finally, a single MAg layer (with 2 output channels) is applied, using a linear activation to produce the desired output mesh (the displacement mesh must have the same structure as the input mesh of forces). It is worth noting that analogous architectures of CNN U-Net networks are similar, with the only distinction being the use of convolution layers in place of MAg layers and CNN U-Net max poolings instead of graph poolings.

As demonstrated in Table~\ref{tab: architectures}, both 2D L-shape and 3D beam examples have been modeled utilizing both MAgNET and CNN U-Net architectures. These networks were designed to have a similar number of trainable parameters, thus facilitating a fair comparison of their fitting capabilities. The number of parameters in both types of networks is controlled by having a higher number of channels in the CNN U-Net architecture compared to its corresponding MAgNET architecture. This difference in the number of channels is attributed to the convolution operators in the CNN architecture sharing parameters across a layer, which may necessitate a larger number of channels to ensure an optimal fit, while the aggregation operators in the MAg layer use individual weights per aggregation window, allowing for more flexible fitting across the mesh with a smaller number of channels. However, caution must be exercised when selecting the number of channels, as setting it too low can result in increased prediction errors (as seen in Figure~\ref{fig: Lconvergence}).

The number of neural network levels (pooling operations) and the size of con\-vo\-lu\-tion/aggre\-ga\-tion windows have been adjusted on a case-by-case basis to obtain the desired fitting capabilities while keeping the number of trainable parameters low and comparable between the respective CNN U-Net and MAgNET models. The fitting capabilities heavily depend on the successful propagation of information from the input throughout the network. This can be compromised when the number of poolings or the window size is too small, as explained in Section~\ref{sec: information-passing interpretation}. For this reason, a larger number of pooling operations is used for mesh graphs with larger diameters (e.g., the 3D cases in Table~\ref{tab: architectures}). Additionally, in the case of MAgNET models, a global optimization of graph pooling operations is performed to reduce the number of nodes at the coarsest level. In this optimization, Algorithm~\ref{alg:adjacency algo} is run 1000 times with different random seeds, and the case with the least number of nodes at the lowest level is selected.

\begin{figure}[h]
     \centering
     \includegraphics[width=0.7\textwidth]{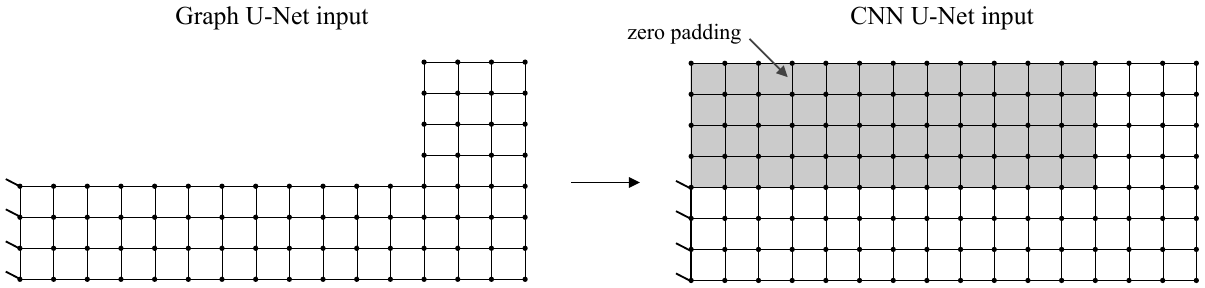}
     \caption{Zero-padding is applied to make the L-shape topology compatible with the CNN framework. The additional nodal values for inputs (forces) and outputs (displacements) are set as zero vectors.}
     \label{fig: L_padding}
\end{figure}

\emph{Remark:} Note that the example presented in Fig.~\ref{fig: benchmark examples schematics}(a) utilizes a non-structured mesh. As such, it can not be directly used by the CNN U-Net model, and an additional preprocessing step needs to be done to make the input and output meshes structured. In this case, we apply zero padding to convert the L-shape mesh into a structured mesh, see Figure~\ref{fig: L_padding}, which is then used for training with the CNN U-Net architecture. We do not need to do this preprocessing step for the MAgNET architecture.  

\begin{figure}[t!h]
     \centering
     \includegraphics[width=0.6\textwidth]{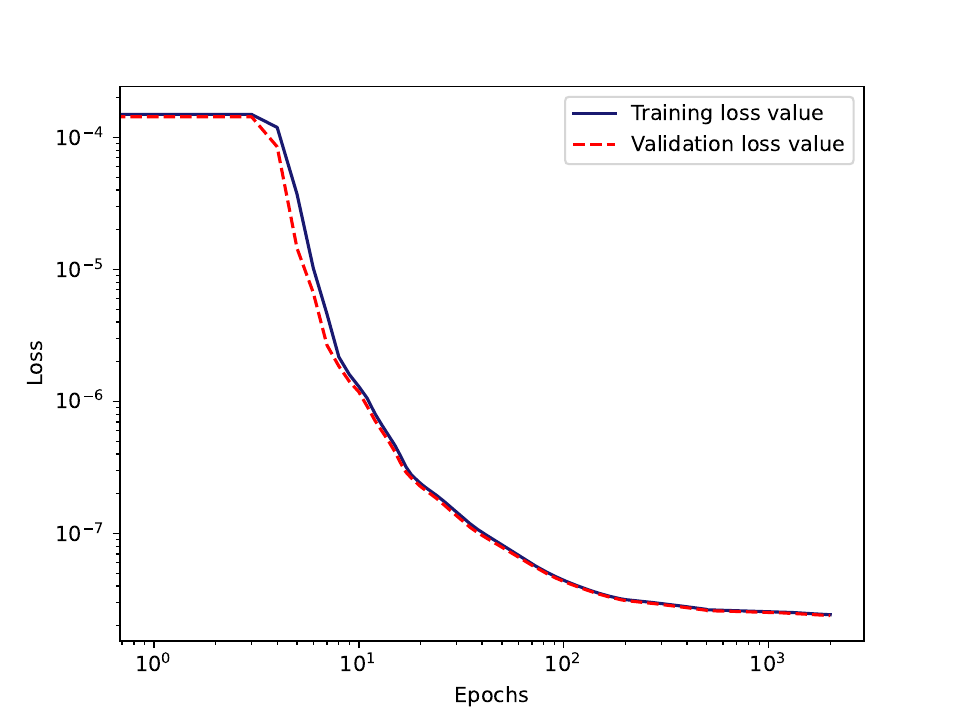}
     \caption{Training loss curve for the 3D breast MAgNET model.}
     \label{fig: breast_loss}
\end{figure}

The models presented in Table~\ref{tab: architectures} were trained by minimizing the loss function, as described in Equation~\ref{eq:lossDeterm}, using the datasets introduced in Section~\ref{sec: data_generation_FEM}. The Adam optimizer, an extension of the stochastic gradient descent algorithm, was used for this purpose. A mini-batch size of 4 and an initial learning rate of $1 \times 10^{-4}$, with a linear decay to $1 \times 10^{-6}$ during training, were employed. The number of epochs (i.e., iterations of the Adam optimizer) was manually tailored on a case-by-case basis to achieve low values of the loss function. An example of the training loss is provided in Figure~\ref{fig: breast_loss}. The network trainings were conducted using TensorFlow on a Tesla V100-SXM2 GPU at the HPC facilities of the University of Luxembourg, see \citep{ULHPC}.

\subsection{Cross validation of CNN U-Net and MAgNET predictions}\label{sec: cross_validation_of_NN_models}



\begin{figure}[!h]
     \centering
     \subfloat[]{\includegraphics[width=0.31\textwidth]{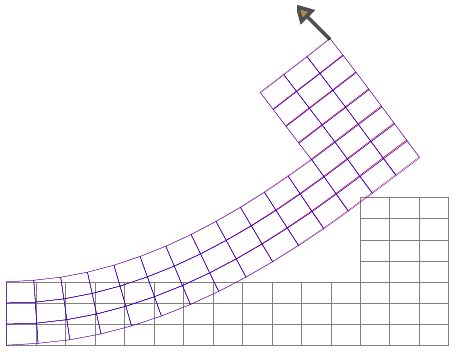}\label{fig: mesh L shape}}\hspace{0.015\textwidth}
     \subfloat[]{\includegraphics[width=0.31\textwidth]{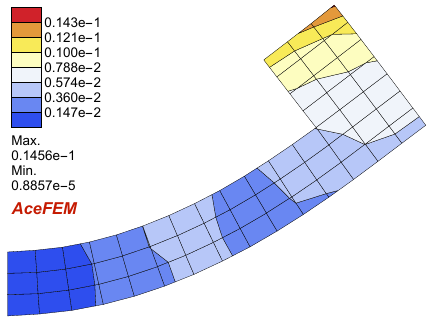}\label{fig: L-GUNet-error}}\hspace{0.015\textwidth}
     \subfloat[]{\includegraphics[width=0.31\textwidth]{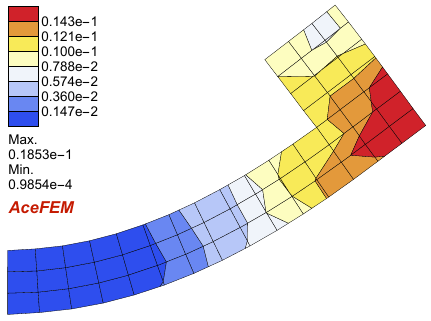}\label{fig: L-CNN-error}}
    \caption{Deformation of 2D L-shape under point load (-0.93, 0.91)N on the corner node (a) Deformed mesh predicted using MAgNET (blue), for comparison FEM solution is presented (red) (b) $L_2$ error of nodal displacements between MAgNET and FEM solution. The relative error for the corner node displacement using MAgNET is 0.5\% (c) $L_2$ error of nodal displacements between CNN U-Net and FEM solution. The relative error for the corner node displacement using CNN is 0.3\%.} 
    \label{fig: 2D L-shape prediction}
\end{figure}

\begin{figure}[!h]
     \centering
     \subfloat[]{\stackinset{r}{-0.015\textwidth}{t}{-.0475\textwidth}{\includegraphics[width=0.07\textwidth]{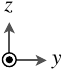}}{\includegraphics[width=0.33\textwidth]{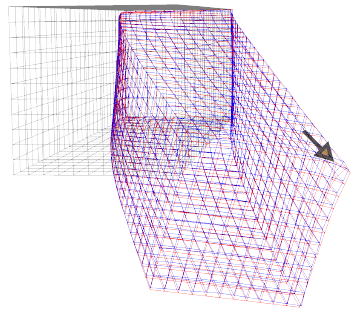}\label{fig: mesh 3d}}\hspace{0.035\textwidth}}
      \subfloat[]{\stackinset{r}{-.05\textwidth}{t}{-.0475\textwidth}{\includegraphics[width=0.07\textwidth]{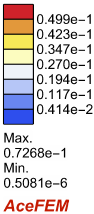}}{\includegraphics[width=0.215 \textwidth]{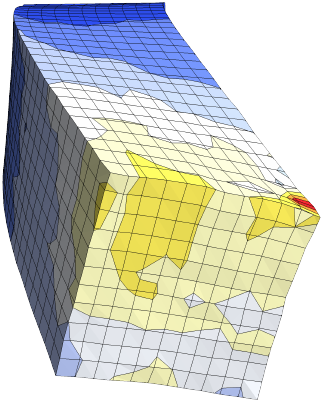}\label{fig: 3d-GUNet-error}}}\hspace{0.08\textwidth}
      \subfloat[]{\stackinset{r}{-.05\textwidth}{t}{-.0475\textwidth}{\includegraphics[width=0.07\textwidth]{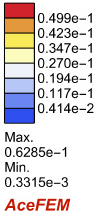}}{\includegraphics[width=0.215 \textwidth]{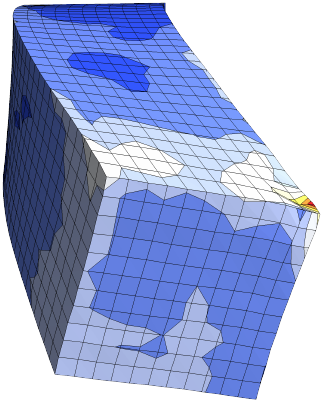}\label{fig: 3d-CNN-error}}}
    \caption{Deformation of the 3D beam under point load (-1.75,1.31,-1.7)N on the second last node (a) Deformed mesh predicted using MAgNET (blue), for comparison FEM solution is presented (red) and undeformed mesh is represented by gray (b) $L_2$ error of nodal displacements between MAgNET and FEM solution. The relative error in predicting displacement of the node of application of load using MAgNET is 4.4\% (c) $L_2$ error of nodal displacements between CNN U-Net and FEM solution. The relative error in predicting displacement of the node of application of load using CNN is 3.0\%.} 
    \label{fig: 3D beam prediction}
\end{figure}

We are going to compare the predictions of MAgNET and CNN U-Net models for two problems with structured inputs/outputs that were introduced in Figure~\ref{fig: benchmark examples schematics}(a,b). Let us look at the individual examples with the highest nodal displacement magnitudes. In the 2D L-shape example, shown in Figure~\ref{fig: 2D L-shape prediction}a, MAgNET predictions visually coincide with the reference FEM solution very well. This is quantitatively shown in Figures~\ref{fig: 2D L-shape prediction}b and~\ref{fig: 2D L-shape prediction}c, where the the $L_2$ error field
\begin{equation}
    \text{err}(\bfm{X})=||u_{\text{FEM}}(\bfm{X})-u_{\text{pred}}(\bfm{X})||_2
    \label{eq: L2ErrorNorm}
\end{equation}
is presented for MAgNET and CNN U-Net, respectively, demonstrating low level of errors for both models. A similar tendency can be observed in the 3D beam case shown in Figure~\ref{fig: 3D beam prediction}. Here, although the level of errors is relatively a bit higher than in the 2D example, the MAgNET and CNN U-Net perform similarly, which proves good capabilities of the proposed MAgNET model as compared to the CNN U-Net model.


In the following, we will analyze and compare the performance of both models for all cases in the test datasets. For that purpose, we need aggregated error metrics. As an error metric for a single test example, we use the mean absolute error,
 \begin{equation}
     e_m = e(\bfm{f}_m,\bfm{u}_m) = \frac{1}{\mathcal{F}}\sum_{i=1}^{\mathcal{F}}{|\mathcal{G}(\bfm{f}_m)^{i} - \bfm{u}_m^{i}|},
     \label{eq:singletest}
 \end{equation}
 where the force-displacement pair $(\bfm{f}_m,\bfm{u}_m)$ is an element of the test dataset 
 \begin{equation}
    \mathcal{D}_{\text{te}}=\{(\bfm{f}_{M_{\text{tr}}+1},\bfm{u}_{M_{\text{tr}}+1}),...,(\bfm{f}_{M_{\text{tr}}+M_{\text{te}}},\bfm{u}_{M_{\text{tr}}+M_{\text{te}}})\},
 \end{equation}
and $\mathcal{F}$ is the number of dofs of the mesh. The metric $e_m$ gives us the notion of error between an expected finite element solution, $\bfm{u}_m$, and the prediction of the neural network, $\mathcal{G}(\bfm{f}_m)$. To analyze the overall quality of fitting, we define a single error metric over the entire test set as the average mean absolute error 
 \begin{equation}
     \Bar{e} =\frac{1}{M_{\text{te}}} \sum_{m=M_{\text{tr}}+1}^{M_{\text{tr}}+M_{\text{te}}} e_{m},
\label{eq:error_metric_det}
 \end{equation}
with the corrected sample standard deviation (standard deviation of averaged errors) defined as
 \begin{equation}
     \sigma(e) = \sqrt{\frac{1}{M_{\text{te}}-1}\sum_{m=M_{\text{tr}}+1}^{M_{\text{tr}}+M_{\text{te}}} \left(e_{m} - \Bar{e} \right)^2}.
 \end{equation}
Finally, in addition to that, we also use the maximum error per degree of freedom over the entire test set
 \begin{equation}
    e_{\text{max}}=\max_{m, i}|\mathcal{G}(\bfm{f}_m)^{i} - \bfm{u}_m^{i}|.
    \label{eq:error_metric_max}
 \end{equation}


 
 


\begin{table}[!th]
\small
\begin{center}
 \begin{tabular}{l| c | c | c | c  } 
 Example & $M_{\text{te}}$ & $\Bar{e}$ [m] & $\sigma(e)~[\text{m}]$& $e_{\text{max}}~[\text{m}]$  \\
 \hline
2D L-shape (MAgNET) & \multirow{2}{*}{200} & 0.5 E-3 & 0.2 E-3 &  1.1 E-2 \\ 
 2D L-shape (CNN U-Net) &  & 0.7 E-3 & 0.6 E-3 &  1.8 E-2 \\[0.3em]
 3D beam (MAgNET) & \multirow{2}{*}{1782}  & 0.8 E-3 & 0.7 E-3 & 7.7 E-2  \\
 3D beam (CNN U-Net) & & 0.7 E-3 & 0.5 E-3 &  5.4 E-2  
\end{tabular}
\end{center}
\caption{Error metrics for the structured mesh examples. $M_{\text{te}}$ stands for the number of test examples, and $\Bar{e}$, $\sigma(e),e_{\text{max}}$ are error metrics defined by Equations~(\ref{eq:error_metric_det})-(\ref{eq:error_metric_max}).} 
\label{tab: stuctured metrics}
\end{table}


\begin{figure}[!ht]
     \centering
     \subfloat[]{\includegraphics[width=0.5\textwidth]{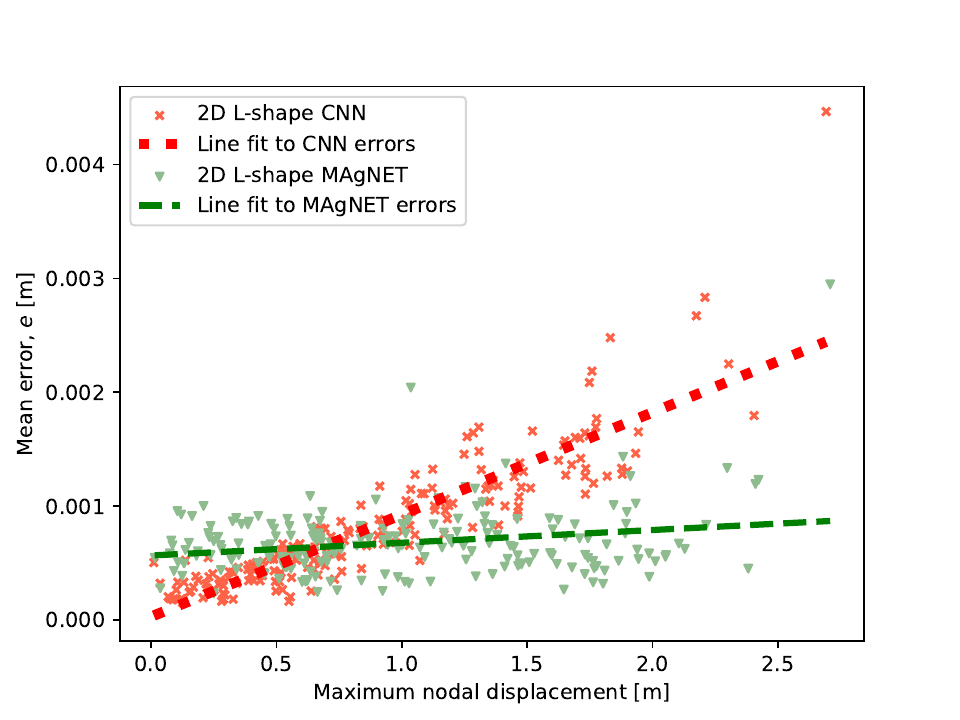}}
     \subfloat[]{\includegraphics[width=0.5\textwidth]{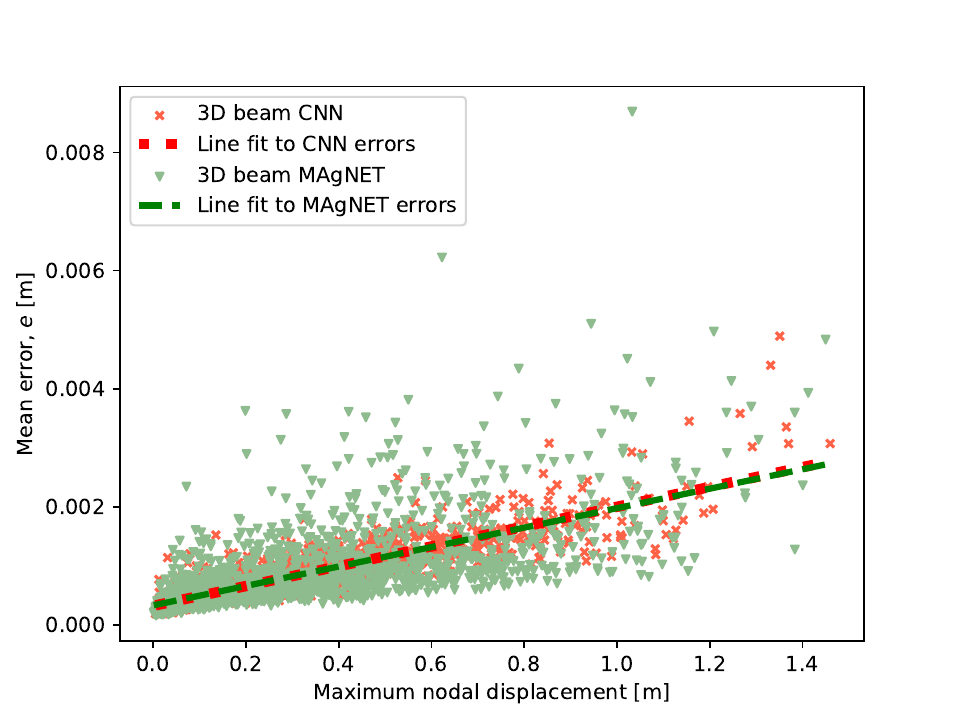}}
      \caption{Mean absolute errors (see Equation~(\ref{eq:singletest})) as a function of maximum nodal displacements for all test examples for 2D L-shape and 3D beam cases for CNN U-NET and MAgNET nerworks.} 
     \label{fig: avg_error_structure}
\end{figure}

The aggregated error metrics for the entire test sets of structured mesh examples obtained using MAgNET and CNN approaches are summarised in Table~\ref{tab: stuctured metrics}. The first observation is that both MAgNET and CNN models exhibit similar prediction accuracy, demonstrating that the MAgNET architecture can achieve a comparable predictive capacity to the CNN U-Net architecture for a similar number of trainable parameters. The prediction errors, with respect to a characteristic length of 1m, fall below 0.1\% for the average mean absolute error (Equation~(\ref{eq:error_metric_det})), which is a promising result given the presence of geometric and constitutive nonlinearities. Additionally, we analyze the performance of the MAgNET model as a function of the maximum nodal displacement per test example. This dependency is visualized in Figure~\ref{fig: avg_error_structure} for both benchmark examples. Although there is a general trend of increased errors for larger maximum displacement magnitudes, the sensitivity is low, and the errors remain small (the regression lines are $e(d) \propto 1.0\cdot{}d\cdot{}10^{-4}$ (2D L-shape) and $e(d) \propto 1.6\cdot{}d\cdot{}10^{-3}$ (3D beam) for MAgNET and $e(d) \propto 7.0\cdot{}d\cdot{}10^{-4}$ (2D L-shape) and $e(d) \propto 1.6\cdot{}d\cdot{}10^{-3}$ (3D beam)) for the CNN U-NET case.

\begin{figure}[!ht]
    \centering
    \subfloat[]{\includegraphics[width=0.49\textwidth]{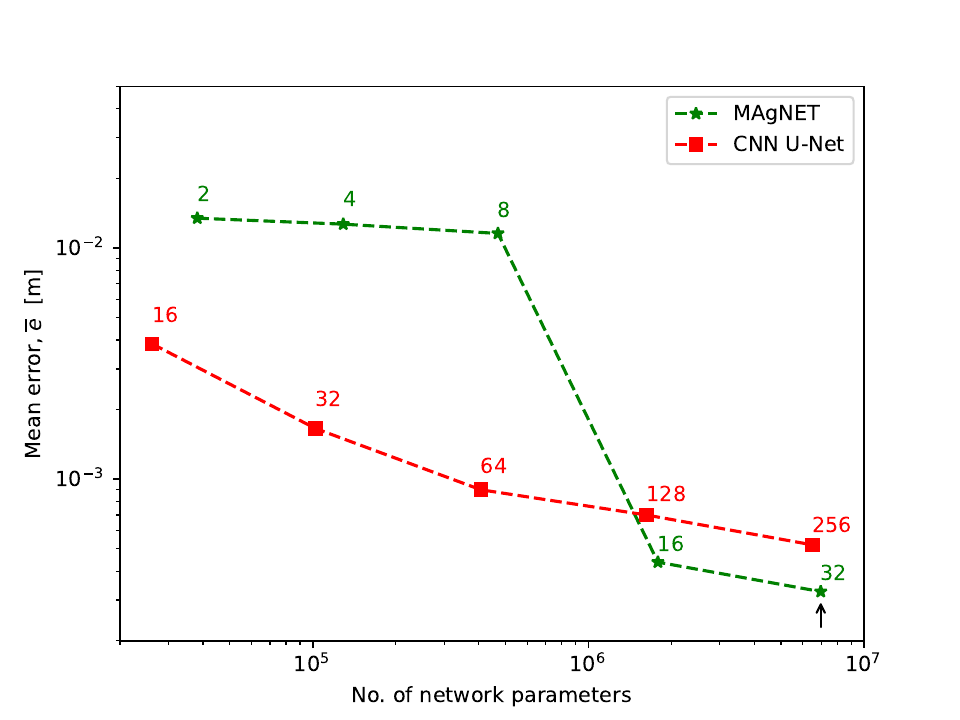}\tikzmark{left}}
    \subfloat[]{\includegraphics[width=0.49\textwidth]{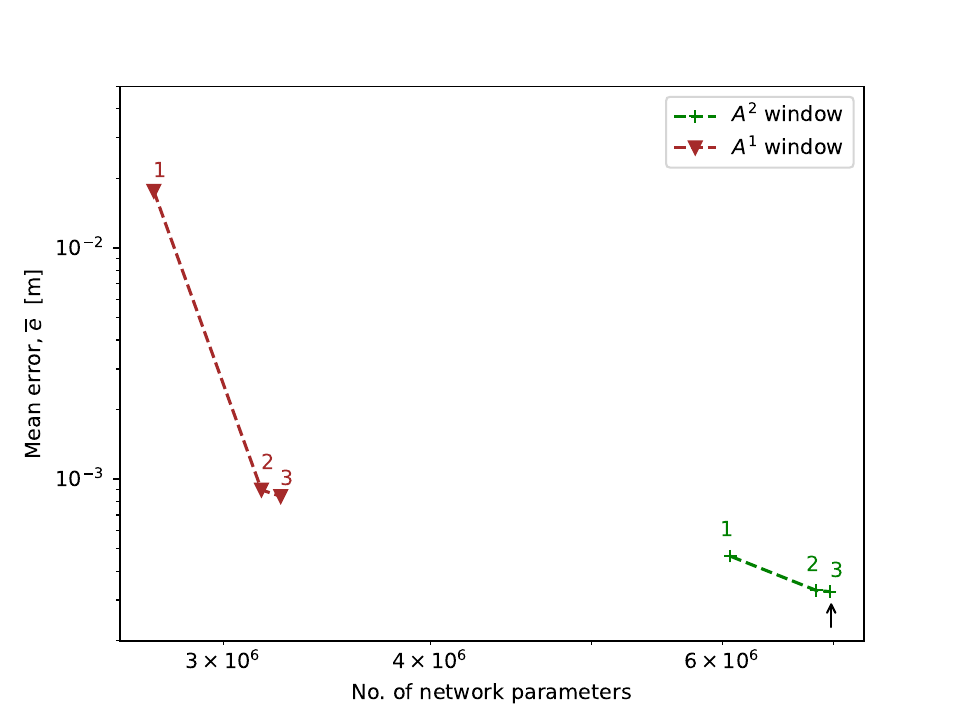}\tikzmark{right}}
    \caption{\revised{Average mean error over the test set for the L-shape case for different network architectures. The reference case for the MAgNET is marked with black arrows in both plots, and represents the architecture of $A^2$ window size, three pooling layers (four levels), and 32 channels. (a) The effect of changing the number of channels used for MAgNET and CNN U-Net architectures (number of channels are marked as numbers in the plot). (b) The effect of reducing the MAg window size and number of pooling layers for MAgNET architecture with 32 channels at each level (number of pooling layers are marked as numbers in the plot).}}
    \label{fig: Lconvergence}

    
\end{figure}


\revised{The increasing sizes of input/output intensify the challenge in non-convex optimization, necessitating more training data and additional epochs for convergence. This trend is universally observed across all deep learning-based surrogate techniques and relates to the well-known concept of the 'curse of dimensionality' in machine learning applications. As discussed in Section~\ref{sec: NN_implementation_and_training}, the complexities of the MAgNET and CNN U-Net architectures are tailored to specific mesh structures and sizes, ensuring effective predictive capabilities while maintaining a comparable number of training parameters. Below, we conduct a simple ablation study to determine the extent to which accuracy declines when simplifying the DNN architectures.}

\revised{In the first ablation test we reduce the number of channels in MAg and CNN layers. As explained in Section~\ref{sec: Mag layer}, the number of channels modulates the model capacity to capture non-linearities in the underlying data}. Importantly, the convolution windows are shareable in CNN architectures, whereas the aggregation windows in MAg architectures are independent. To this end, we expect CNN networks to require more channels than their respective MAgNET networks to achieve the same level of accuracy. We used this fact when designing CNN and MAgNET architectures in Section~\ref{sec: NN_implementation_and_training}. To verify this hypothesis and demonstrate this effect, we trained five MAgNET and five CNN U-Net networks on the L-shape dataset with different numbers of channels. In all analyzed cases, we used 4-level MAgNET and 3-level CNN U-Net architectures, with two MAg/Conv layers per level, and with a constant number of channels at all levels. Figure~\ref{fig: Lconvergence}(a) shows that there is indeed a strong dependency of accuracy on the number of channels for both analyzed network architectures. For a comparable number of trainable parameters, CNN U-Nets can use more channels than their respective MAgNETs, providing them with comparable predictive accuracy. We can also observe that too few channels significantly reduce the fitting capabilities of both networks, with a step jump between the 8- and 16-channel case for MAgNET. For the two largest cases (16 and 32 channels for MAgNET and 128 and 256 channels for CNN U-Net), the accuracy of both architectures is comparable.

\revised{In the second ablation test, we modulate two other important hyperparameters of MAgNET architectures: the MAg layer window size and the number of pooling layers. Again, we utilize the L-shape dataset, with the reference MAgNET architecture characterized by a window size of $A^2$, four levels (three pooling operations), and 32 channels at each level. Figure~\ref{fig: Lconvergence}(b) demonstrates a general trend where the model's simplification leads to a drop in accuracy. This is observable both when reducing the window size from $A^2$ to $A^1$ and when decreasing the number of levels (pooling operations). This reduced capacity of the simplified models can be explained by the overall reduced number of parameters, but also by the less intensive exchange of information across the graph. In particular, a significant drop in accuracy is observed for the single-pooling $A^1$ case. This can be attributed to the inability of this oversimplified network to fully exchange the nodal input information, confirming the effect discussed in more detail in Section~\ref{sec: information-passing interpretation}. 
}




\revised{\textit{Remark}: 
The high accuracy of the proposed surrogate models is accompanied by very rapid prediction times compared to the respective high-fidelity FEM predictions. For example, in the 3D beam case, the nonlinear FEM solution required over 3~seconds for some test examples, while MAgNET could make predictions in just 0.18~seconds. Moreover, while MAgNET maintains consistent prediction times regardless of the input force, the time for the FEM solution increases with larger load cases. This increase is due to the use of an iterative solver and an adaptive load-stepping scheme, which are necessary to address convergence issues in highly nonlinear problems. A more comprehensive study on how MAgNET's computational efficiency compares with nonlinear FEM, as well as with CNN-based and attention-based neural networks, is provided in~\cite{Deshpande2023}. }

\subsection{Predictions of MAgNET for general (unstructured) meshes}
\label{sec: predictions_MAgNET_unstructured}

In Section~\ref{sec: cross_validation_of_NN_models}, we demonstrated that the MAgNET architectures can achieve very good predictive capabilities for structured mesh cases, which was also cross-validated against respective CNN U-Net architectures. In this section, we aim to show that the high prediction accuracy of MAgNET can also be expected for unstructured mesh cases, which is the central point of the results section. We consider two cases: the first one is deformation under the application of point loads (the 2D beam with hole case), similar to the case of structured examples, and the second is deformation under body forces (the 3D breast case, inspired from \citep{LAVIGNE2023115889}).

\begin{figure}[t!h]
     \centering
     \subfloat[]{\includegraphics[width=0.45\textwidth]{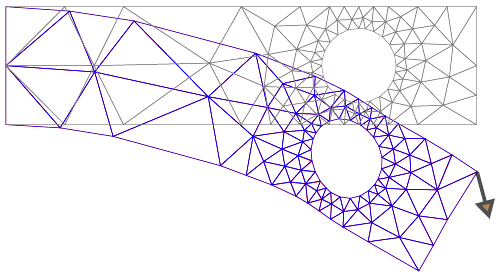}}\hspace{0.04\textwidth}
     \subfloat[]{\includegraphics[width=0.50\textwidth]{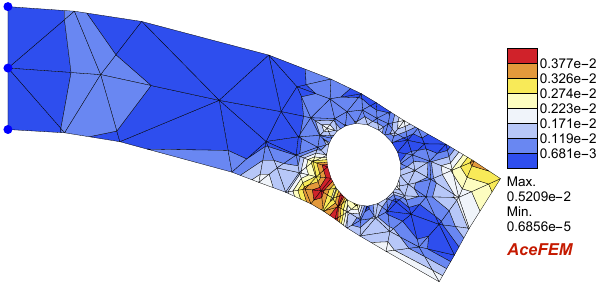}\label{fig: error_multiple}}\\
    \subfloat[]{\includegraphics[width=0.45\textwidth]{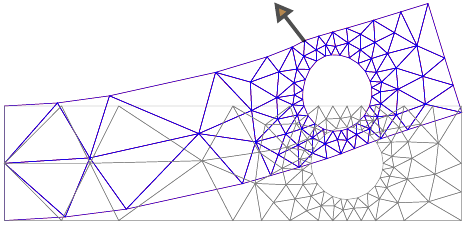}}\hspace{0.04\textwidth}
    \subfloat[]{\includegraphics[width=0.50\textwidth]{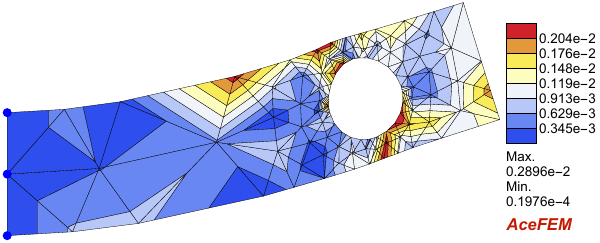}\label{fig: error_multiple2}}
 \caption{Deformation of the 2D beam under two different point loads (upper case: ($1.28,-4.43$)N, lower case: ($-3.38,4.04$)N). (a)\&(c) Deformed meshes computed using MAgNET (blue) and FEM (red), with the undeformed configuration (gray).
 (b)\&(d) $L_2$ error of nodal displacements between MAgNET and FEM solutions.}
    \label{fig: 2D beam with hole}
\end{figure}
\begin{figure}[h]
     \centering
     \subfloat[]{\includegraphics[height=0.41\textwidth]{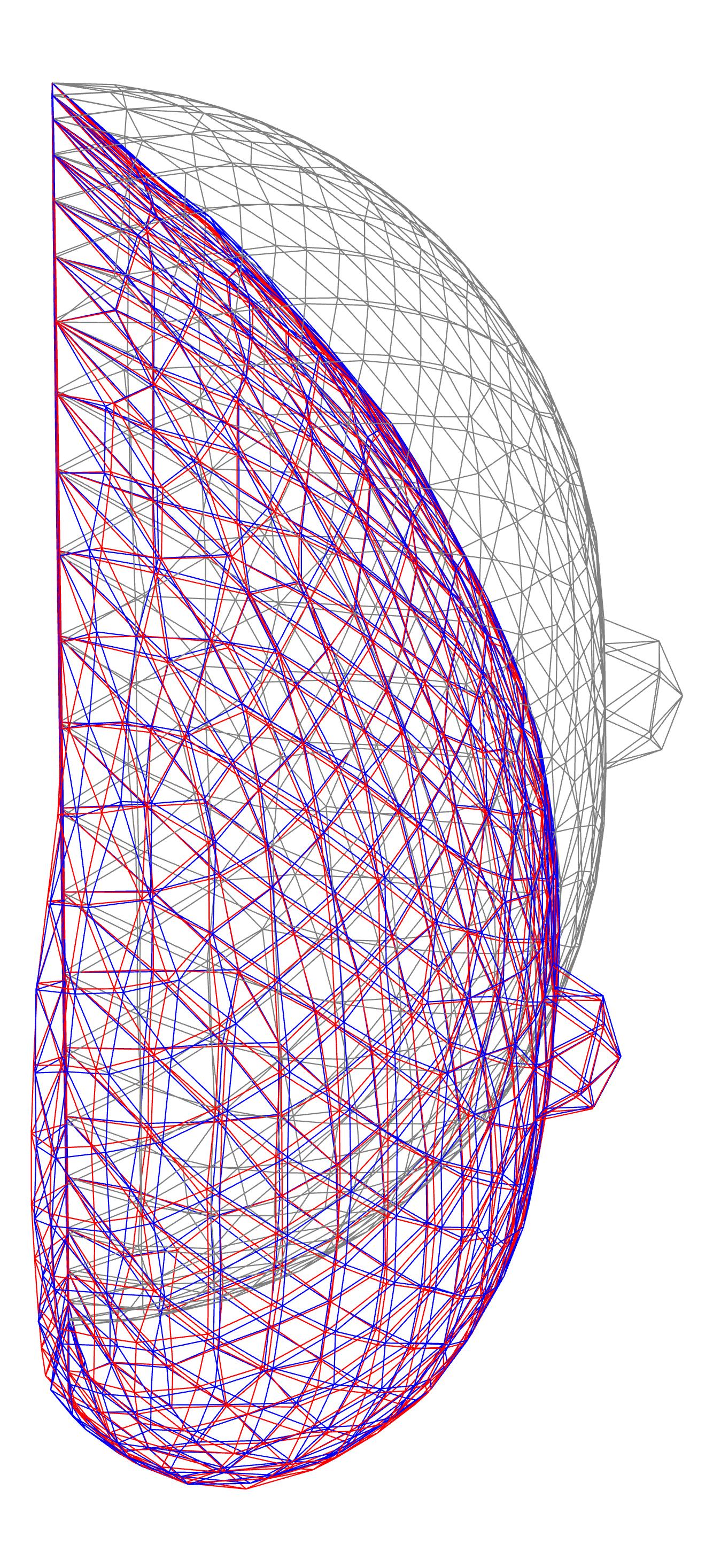}\label{breast_mesh}}
     \hspace{0.05\textwidth}
     \subfloat[]{\stackinset{r}{0.28\textwidth}{t}{0.005\textwidth}{\includegraphics[width=0.05\textwidth]{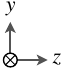}}{\includegraphics[height=0.4\textwidth]{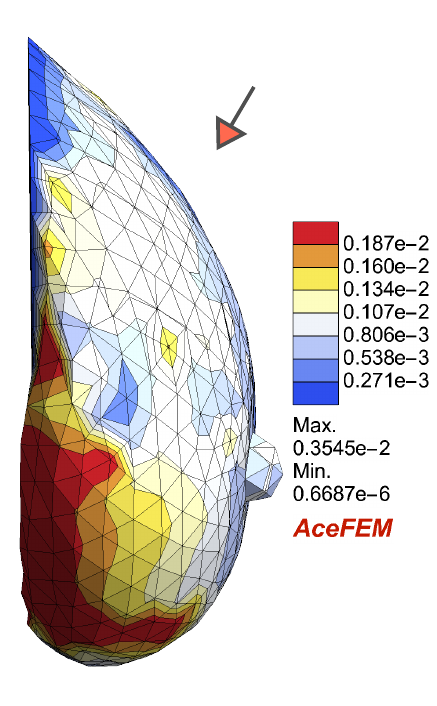}\hspace{0.01\textwidth}\vspace{0.1\textwidth}\label{breast_error}}}
     \hspace{0.02\textwidth}
     \subfloat[]{\includegraphics[height=0.37\textwidth]{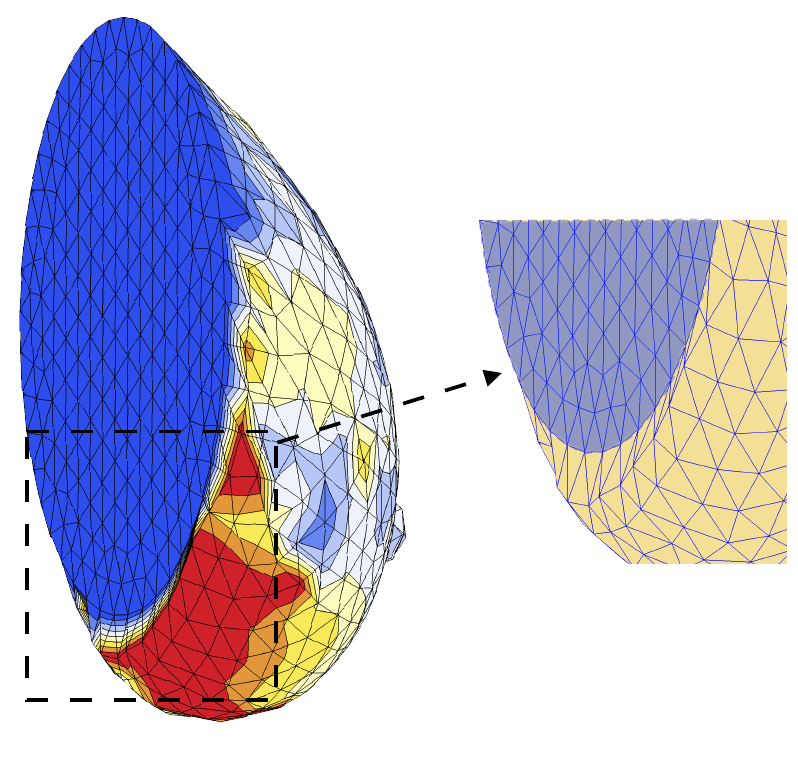}}
     \caption{Deformation of the 3D breast geometry with force density of ($-5.94, -5.23, -2.56$)~$\text{N}/\text{kg}$. (a) Deformed meshes computed using MAgNET (blue) and FEM (red), with the undeformed configuration (gray). 
     (b) $L_2$ error of nodal displacements between MAgNET and FEM solutions. (c) Titled view of the figure(b), MAgNET efficiently captures fixed boundary and nearby high non-linear deformations by learning implicitly from the data.} 
     \label{fig: 3d_breast_viz}
\end{figure}

Let us first analyze individual examples. In Fig.~\ref{fig: 2D beam with hole} we present two particular loading cases of the 2D beam with hole. One of them is loaded at the tip, featuring the highest nodal displacement magnitude of all test cases, and the other one is loaded close to the hole, representing high local distortions. Similarly, for the three-dimensional problem, in Fig.~\ref{fig: 3d_breast_viz} we show the case featuring the highest nodal displacement magnitude of all test cases. In all mentioned examples we can observe overall good accuracy when visually comparing MAgNET predictions with the respective FEM solutions. This can also be checked quantitatively by analyzing maximum displacement errors. In the cases of the 2D beam with hole, those errors are 1.4\% and below when related to the characteristic length of~1m. In the case of 3D breast geometry, such relative maximum error is higher, reaching almost 3.1\% (related to the breast diameter of 0.16m). Despite this fact, we can observe that high local shape distortions are very well recovered. This property is more emphasized in Fig.~\ref{fig: 3d_breast_viz}c where one can additionally observe that also the Dirichlet boundary conditions are very well predicted, even though they were only introduced implicitly by training data.

\begin{figure}[!ht]
     \centering
     \subfloat[]{\includegraphics[width=0.5\textwidth]{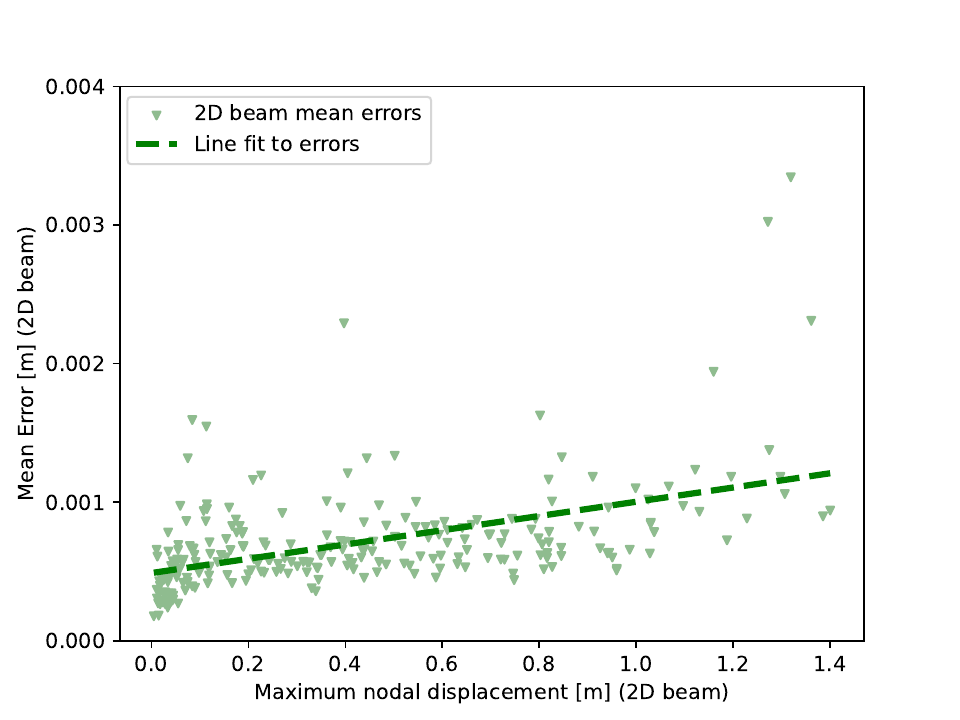}}
     \subfloat[]{\includegraphics[width=0.5\textwidth]{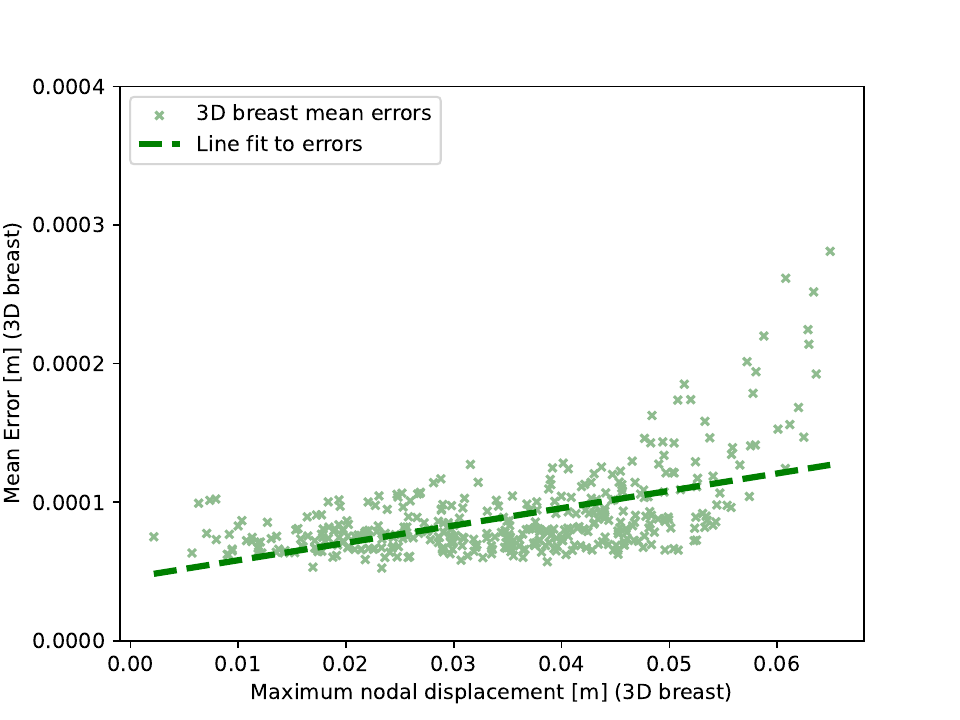}}
       \caption{Mean absolute errors (see Equation~(\ref{eq:singletest})) as a function of maximum nodal displacements for all test examples (with unstructured meshes) predicted using MAgNET for (a) 2D beam with hole (b) 3D breast case.}
     \label{fig: avg_error_unstructured}
\end{figure}

The aggregated error metrics for the entire test sets are provided in Table~\ref{tab: unstructured_metrics}. The maximum displacement errors over all test cases, $e_{\text{max}}$, are at the levels observed for particular cases in Figures~\ref{fig: 2D beam with hole} and~\ref{fig: 3d_breast_viz}. At the same time, the average mean errors, $\Bar{e}$, are at least an order of magnitude lower, which suggests that the errors close to maximum levels are not that often. The average mean errors are further analyzed in a case-by-case manner in Fig.~\ref{fig: avg_error_unstructured}, which is analogous to the analysis done for the structured cases in Fig.~\ref{fig: avg_error_structure}. Again, we plot the mean error $e$, of each test example as a function of the maximum nodal displacement. The regression lines $e(d) \propto 5.0 \cdot d \cdot 10^{-4}$ (2D-beam) and  $e(d) \propto 8.0 \cdot d \cdot 10^{-4}$ (3D Breast) show low sensitivity of the MAgNET predictions to displacement magnitudes.

\begin{table}[h]
\small
\begin{center}
 \begin{tabular}{l | c | c | c | c   } 
Example & $M_{\text{te}}$ & $\Bar{e}$ [m] & $\sigma(e)~[\text{m}]$& $e_{\text{max}}~[\text{m}]$ \\ [0.5ex] 
 \hline
 2D beam (hole) & 240 & 0.7 E-3 & 0.4 E-3 &  1.4 E-2 \\
3D breast & 400 & 8.9 E-5 & 3.1 E-5 &  5.1 E-3
\end{tabular}
\end{center}
\caption{Error metrics for the unstructured mesh examples. $M_{\text{te}}$ stands for the number of test examples, and $\Bar{e}$, $\sigma(e),e_{\text{max}}$ are error metrics defined in Section~\ref{sec: cross_validation_of_NN_models}. 
}
\label{tab: unstructured_metrics}
\end{table}

\begin{figure}[h]
     \centering
     \subfloat[]{\stackinset{r}{0.22\textwidth}{t}{0.0885\textwidth}{\includegraphics[width=0.05\textwidth]{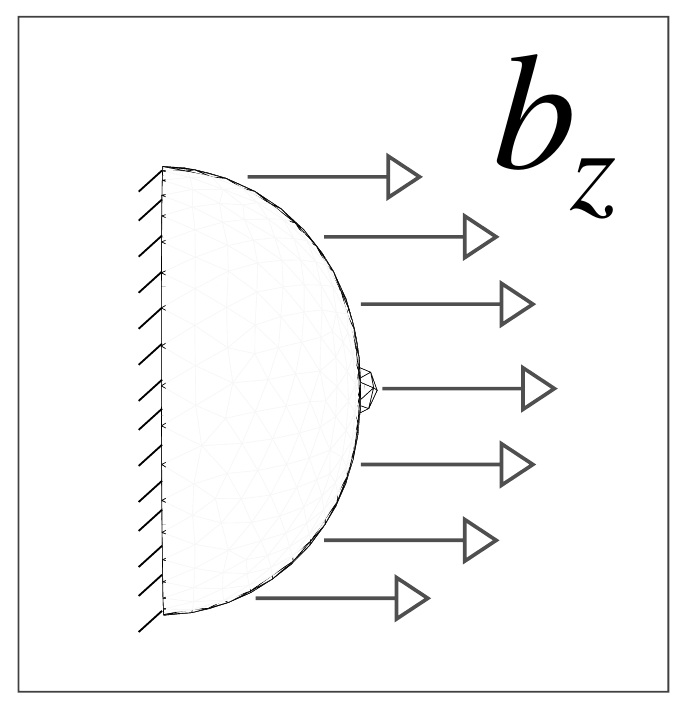}}{\includegraphics[width=0.5 \textwidth]{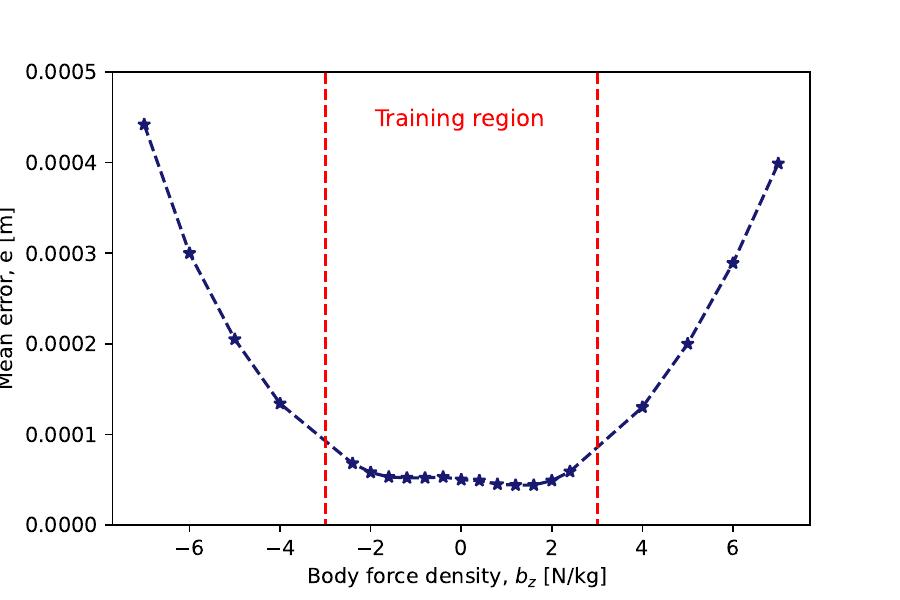}\label{errorvsbz}}}
     \subfloat[]{\includegraphics[width=0.22\textwidth]{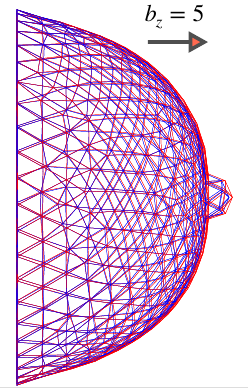}\label{bz5}}\hspace{0.03\textwidth}
     \subfloat[]{\includegraphics[width=0.22\textwidth]{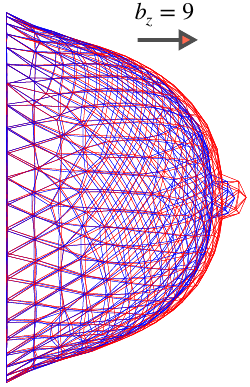}\label{bz9}}
     \caption{3D Breast deformation under horizontal body force densities, $(0,0,b_{\text{z}})\,\text{N}/\text{kg}$. (a) Mean absolute error for testing cases in interpolated and extrapolated regions. The error increases rapidly in the extrapolated region while it remains low in the training (interpolated) region. (b)\&(c) Visualisation of deformed meshes for force densities outside the training region computed using MAgNET (blue) and FEM solution (red). 
     } 
     \label{fig: breast_extrapolate}
\end{figure}

Above, we have demonstrated a good prediction accuracy of MAgNET within the test dataset (which is located in the interpolated domain). However, it is well known that this accuracy can gradually deteriorate when moving to the extrapolated region, see, e.g., \cite{DESHPANDE2022115307}. We are going to study this effect for MAgNET for a particular case that is based on the 3D breast geometry. As described in the Table~\ref{tab:datasets}, during the training, the $b_{\text{z}}$ component of body force density is varied from -3 to 3 $\text{N}/\text{kg}$ only. At the inference time, we applied $b_{\text{z}}$ from -7 to 7 $\text{N}/\text{kg}$ (keeping other components 0) to see how the predictions perform within and outside the training magnitudes. Figure~\ref{errorvsbz} shows that the error is fairly low and is not increasing within the training region and it increases rapidly outside, which confirms this well-known effect.  Figures~\ref{bz5} and~\ref{bz9} show deformed meshes predicted for $b_z$= 5 and $b_z$ = 9 $\text{N}/\text{kg}$, respectively, both outside the training data region. MAgNET is observed to give visually acceptable results although the accuracy of the framework decreases as we move away from the training data.

\subsubsection{A note on physics-informed errors} \label{sec: physics_informed}

The proposed MAgNET framework has only been trained by minimizing the loss function for displacement errors, with no additional explicit information about the underlying physics/mechanics. As demonstrated earlier in this work, such training can provide very good accuracy in terms of predicted displacements. However, this accuracy is not of machine precision. To this end, a natural question arises: how far the displacement errors can violate physics? To answer that question, we are going to analyse some problem-based quantities of interest, such as residuals (balance of forces) or stresses, in comparison to the expected ground-truth results. 

\begin{figure}[!ht]
     \centering
     \subfloat[]{\includegraphics[width=0.47\textwidth]{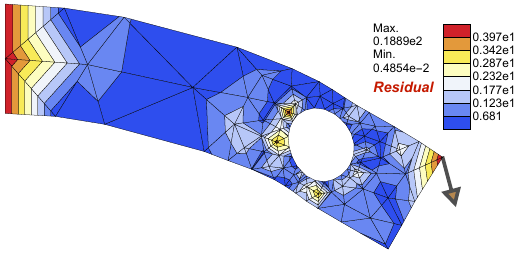}}\hspace{0.04\textwidth}
     \subfloat[]{\includegraphics[width=0.47\textwidth]{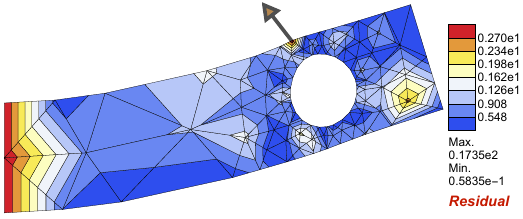}}
      \caption{Nodal residual forces obtained using MAgNET solutions for the examples in Figure~\ref{fig: 2D beam with hole} (plotted on deformed meshes). The relative error for retrieving the total reaction force at the fixed interface is (a) 4.7\% for the first example (b) 0.1\% for the second example. } 
     \label{fig: 2D beam residual}
\end{figure}

In Figure~\ref{fig: 2D beam residual} we show nodal internal residual forces for the 2D beam with hole cases that we introduced earlier (compare Figure~\ref{fig: 2D beam with hole} for respective displacement errors). Ideally, the residual forces should be zero (the balance of forces), except for the boundary condition areas in which they should be exactly opposite to the reaction at the support and the applied external force. However, due to inaccuracies in displacements obtained from the MAgNET model, differences with respect to the ground true residuals can be noted. In Figure~\ref{fig: 2D beam residual}, we can observe the expected high residual forces in the areas where Dirichlet and Neumann boundary conditions are applied, however, also localised residual force spots are present in the fine mesh region around the hole.  The magnitude of those errors in the localised spots can go up to 20\% of the maximal magnitude of applied forces. Also, when more closely analysing the residuals at boundary condition areas, it turns out that they do not fully match the respective FEM residuals. For instance, the relative error in residuals at the support in Figure~\ref{fig: 2D beam residual}a is almost 5\%. When analysing the entire test set, we observed that the mean error in retrieving residuals at the Dirchilet boundary related to the maximum external force is 2.4\%. A similar relative mean error value for retrieving the Neumann boundary residual is 14.6\%. The higher error for Neumann boundary residual is attributed to high local non-linear deformations at the vicinity of the point of application of force. Though the displacement errors provoked by these non-linearities are not so high, they can result in high residual errors through the relatively high magnitude of element stiffness.


\begin{figure}[!ht]
     \centering
     \subfloat[]{\includegraphics[width=0.47\textwidth]{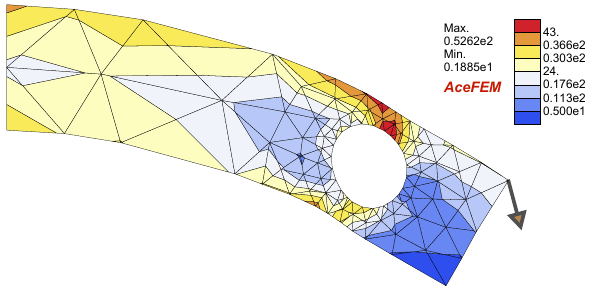}}\hspace{0.04\textwidth}
     \subfloat[]{\includegraphics[width=0.47\textwidth]{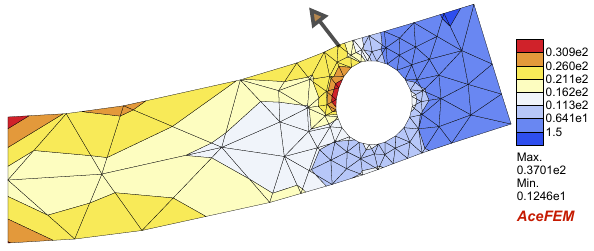}}

     \subfloat[]{\includegraphics[width=0.47\textwidth]{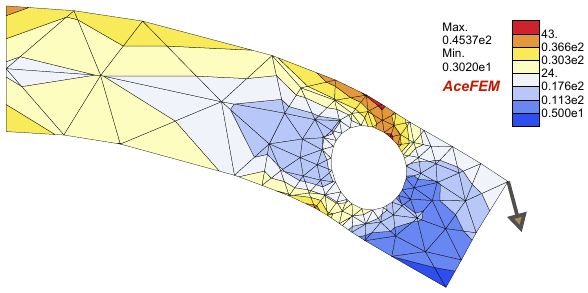}}\hspace{0.04\textwidth}
     \subfloat[]{\includegraphics[width=0.47\textwidth]{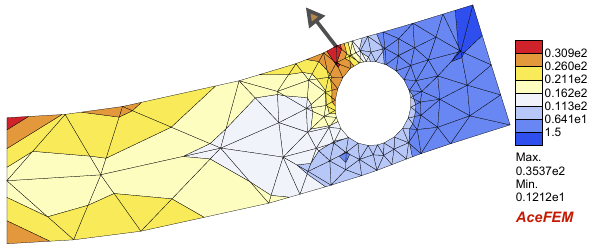}}

     \subfloat[]{\includegraphics[width=0.47\textwidth]{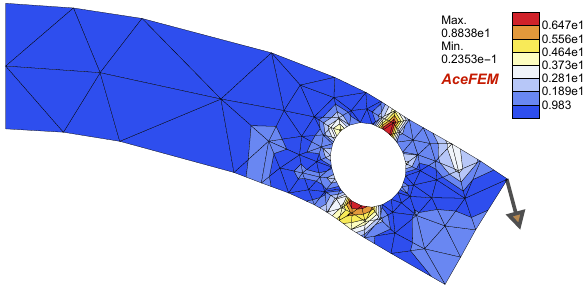}\label{firstmiseserror}}\hspace{0.04\textwidth}
     \subfloat[]{\includegraphics[width=0.47\textwidth]{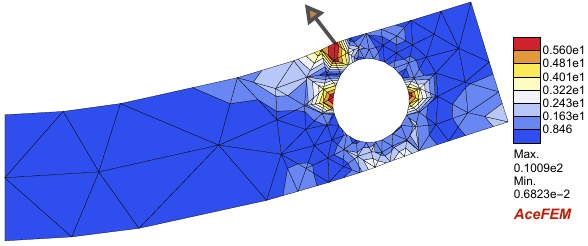}\label{secondmiseserror}}
     
      \caption{Von~Mises stresses obtained for the two examples as in Figure~\ref{fig: 2D beam with hole} using (a)\&(b) MAgNET solution (c)\&(d) FEM solution. In (e)\&(f) the absolute error between the MAgNET and FEM von Mises stresses is shown.}
     \label{fig: 2D beam vom miss}
\end{figure}

The errors observed in Figures~\ref{fig: 2D beam with hole} and~\ref{fig: 2D beam residual} can have a direct impact on some application-dependent quantities of interest. As an example, in Figure~\ref{fig: 2D beam vom miss}, we present the field of von~Mises stresses, which is a commonly used measure of shear stresses. We can observe that the MAgNET solution provides similar profiles of stresses as compared to respective FEM solutions, however, high localised errors are present at the fine mesh region (up to 30\% of the reference FEM maximal von~Mises stresses).

The above mentioned localised errors in residual forces, von Mises stresses and other relevant quantities of interest can be reduced by enriching the loss function with physics-informed terms. For instance, in the context of mesh-based force-displacement data, in \citep{alban} such enrichment has been introduced by scaling individual components of the loss function with the respective computed residual values, which proved to reduce residual errors. In~\citep{mechanicsinformed}, the authors introduced an energy-based approach that provided purely physics-informed training for the Gauss-point stress-strain relationship, which allowed them to satisfy the expected frame indifference. Similar concepts of physics-informed loss functions can be seamlessly integrated into the MAgNET framework, which would convert it into a Physics Informed MAgNET.

\section{\revised{Conclusion and future directions}}\label{sec: Conclusions}


In this work we proposed MAgNET, a novel framework for efficient supervised learning on graph-structured data using geometric deep learning. The framework comprises two neural network operations: MAg and graph pooling/unpooling layers, which together form a graph U-Net architecture capable of learning on large-dimension inputs/outputs. Notably, the MAgNET framework is not restricted to any particular input$\to$output relationship or any specific mesh- or discretization scheme, making it superior to existing convolutional neural network architectures. MAgNET allows for arbitrary non-grid inputs/outputs, meaning it can handle arbitrary meshes and support complex geometries and local mesh refinements, making it suitable for a wide range of engineering applications.

We demonstrated and studied the capabilities of MAgNET in capturing nonlinear relationships in data. In particular, we showed that MAgNET can serve as an efficient surrogate framework for non-linear FEM simulations. For this purpose, we conducted quantitative cross-validation of predictions made by MAgNET and the well-known convolutional U-Net architecture, both of which have been verified against the ground-truth results obtained with FEM. The benchmarks have proven that MAgNET has similar predictive capabilities as CNN U-Net for structured meshes, and it can also be extended to arbitrary meshes while preserving similar accuracy of predictions. 

\revised{There are multiple avenues for enhancing MAgNET's capabilities and applying it beyond the scope of the present work. One promising direction involves directly integrating the underlying physics into the training process. While the current purely data-based MAgNET accurately captures the displacement field, it exhibits increased errors for other physics-related quantities, as discussed in Section~\ref{sec: physics_informed}. Incorporating physics-based components into the loss function could significantly improve MAgNET's performance in these areas. A second potential research direction is applying MAgNET to unsteady problems. Such techniques already exist, for instance, in applications to dynamics \citep{pfaff2021learning, MEISTER2020112628} and in the context of elasto-plasticity \citep{Mozaffar2019, soumi2021arxiv}. We believe some of them could be seamlessly integrated with the MAgNET architecture. A third future research direction is to enhance the prediction accuracy of surrogate models for real-world applications involving online observational data. One approach is to combine any type of DNNs, such as MAgNET, with data assimilation techniques, like Kalman filtering~\citep{kalman}. A fourth potential direction is to transform MAgNET into a probabilistic model. This could be achieved by performing local aggregations with probability distributions instead of discrete weights, similar to our approach with Bayesian CNNs in previous work \citep{DESHPANDE2022115307}. A Bayesian MAgNET would be capable of tracking uncertainties inherent in both the network architecture and real-world data. Finally, we envision applying MAgNET not only to forward problems but also to inverse problems. The MAgNET architecture's ability to handle non-structured meshes could be combined with existing ideas for data-driven calibration of model parameters, as seen, for example, in \citep{PINN}, allowing for a DNN extension of our earlier calibration schemes \citep{LAVIGNE2023115889} to a broader class of systems and models.}


We have made all the codes, datasets, and examples presented in this paper available open-access in the MAgNET repository at \href{https://github.com/saurabhdeshpande93/MAgNET}{https://github.com/saurabhdeshpande93/MAgNET}. Given the generality of MAgNET in supporting arbitrary non-linear relationships and arbitrary discretizations, we believe that the repository will provide a useful surrogate modeling framework for researchers and practitioners in various application areas across disciplines. We see it not only as a ready-to-use machine-learning library but also as a reference point and foundation for future developments and extensions in this emerging direction of research. The generality of MAgNET will enable the community to explore a range of new applications and modeling scenarios.




\vspace{5mm}

\emph{Acknowledgements:}
\begin{wrapfigure}{l}{0.34\textwidth}
    \includegraphics[width=0.3\textwidth]{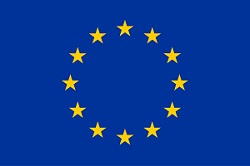}
\end{wrapfigure}
This project has received funding from the European Union’s Horizon 2020 research and innovation programme under the Marie Sklodowska-Curie grant agreement No. 764644. 
Jakub Lengiewicz would like to acknowledge the support from EU Horizon 2020 Marie Sklodowska Curie Individual Fellowship \emph{MOrPhEM} under Grant 800150. St\'ephane Bordas and Jakub Lengiewicz are grateful for the support of the Fonds National de la Recherche Luxembourg FNR grant QuaC C20/MS/14782078. St\'ephane Bordas received funding from the European Union's Horizon 2020 research and innovation programme under grant agreement No 811099 TWINNING Project DRIVEN for the University of Luxembourg. This paper only contains the author's views and the Research Executive Agency and the Commission are not responsible for any use that may be made of the information it contains.

\bibliography{mybibfile}

\end{document}